\documentclass[twoside,11pt]{article}
\pdfoutput=1

\usepackage{geometry}
\usepackage{natbib}
\usepackage{graphicx}
\usepackage{rotating}
\usepackage{amsmath}
\usepackage{amssymb}
\usepackage{theorem}
\newcommand{\BlackBox}{\rule{1.5ex}{1.5ex}}  % end of proof

\newcommand{\cbr}[1]{\left\{#1\right\}}

\newcommand{\intset}[1]{\cbr{1..n}}

\usepackage[colorlinks,pdfpagelabels,plainpages=false]{hyperref}
\usepackage{algorithm}
\usepackage{algorithmic}
\usepackage{rotating}

\usepackage{xcolor}
\definecolor{dark-red}{rgb}{0.4,0.15,0.15}
\definecolor{dark-blue}{rgb}{0.15,0.15,0.4}
\definecolor{medium-blue}{rgb}{0,0,0.5}
\hypersetup{
   colorlinks, linkcolor={dark-blue},
   citecolor={dark-blue}, urlcolor={medium-blue}
}

\usepackage{booktabs}
\usepackage{bm}
\usepackage{amsmath}
\usepackage{amssymb}
\usepackage{amsfonts}
\usepackage{mathtools}
\usepackage{comment}
\usepackage{subfigure}

\newcommand{\mbf}[1]{{\boldsymbol{\mathbf{#1}}}}
\renewcommand{\bm}{\mbf}

\usepackage{color}

\usepackage{array}
\newcolumntype{L}[1]{>{\raggedright\let\newline\\\arraybackslash\hspace{0pt}}m{#1}}
\newcolumntype{C}[1]{>{\centering\let\newline\\\arraybackslash\hspace{0pt}}m{#1}}
\newcolumntype{R}[1]{>{\raggedleft\let\newline\\\arraybackslash\hspace{0pt}}m{#1}}

\usepackage{parskip}

\title{Thoughts on Massively Scalable Gaussian Processes}

\author{
  Andrew Gordon Wilson \\
  Carnegie Mellon University \\
  andrewgw@cs.cmu.edu
  \and
  Christoph Dann \\
  Carnegie Mellon University \\
  cdann@cdann.de
  \and
  Hannes Nickisch \\
  Philips Research Hamburg \\
  hannes@nickisch.org
}

\begin{document}
\date{}  

\maketitle

\begin{abstract} 
\begin{sloppypar}
%\begin{normalsize}

We introduce a framework and early results for massively scalable Gaussian processes (MSGP),
significantly extending the KISS-GP approach of \citet{wilsonnickisch2015}.
The MSGP framework enables the use of 
Gaussian processes (GPs) on billions of datapoints, without requiring distributed inference, or severe assumptions.  In particular,
MSGP reduces the standard $\mathcal{O}(n^3)$ complexity of GP learning and inference to $\mathcal{O}(n)$, and the 
standard $\mathcal{O}(n^2)$ complexity per test point prediction to $\mathcal{O}(1)$.  MSGP involves 
1) decomposing covariance matrices as Kronecker products of Toeplitz matrices approximated by circulant matrices.  
This multi-level circulant approximation allows one to unify the orthogonal computational benefits of fast Kronecker and 
Toeplitz approaches, and is significantly faster than either approach in isolation; 2) local kernel interpolation and inducing points 
to allow for arbitrarily located data inputs, and $\mathcal{O}(1)$ test time predictions; 3) exploiting block-Toeplitz Toeplitz-block structure (BTTB), which enables
fast inference and learning when multidimensional Kronecker structure is not present; and 4) projections of the input space to flexibly model correlated
inputs and high dimensional data.  The ability to handle many ($m \approx n$) inducing points allows for near-exact accuracy
and large scale kernel learning.

%\end{normalsize}
\end{sloppypar}

\end{abstract} 

\section{Introduction}

Every minute of the day, users share hundreds of thousands of pictures, videos, tweets, reviews, and blog posts. 
More than ever before, we have access to massive datasets in almost every area of science and engineering, 
including genomics, robotics, and climate science.  This wealth of information provides an unprecedented opportunity 
to automatically learn rich representations of data, which allows us to greatly improve performance in predictive tasks, 
but also provides a mechanism for scientific discovery.  

Expressive non-parametric methods, such as Gaussian processes (GPs) \citep{rasmussen06}, have great potential for large-scale structure 
discovery; indeed, these methods can be highly flexible, and have an information capacity that grows with the amount of available data. 
However, large data problems are mostly uncharted territory for GPs, which can only be applied to at most a few thousand
training points $n$, due to the $\mathcal{O}(n^3)$ computations and $\mathcal{O}(n^2)$ storage required for inference and learning.

Even more scalable approximate GP approaches, such as inducing point methods \citep{quinonero2005unifying}, typically
require $\mathcal{O}(m^2 n + m^3)$ computations and $\mathcal{O}(m^2+mn)$ storage, for $m$ inducing points, and are hard to apply
massive datasets, containing $n > 10^5$ examples.  Moreover, for computational tractability, these approaches
require $m \ll n$, which can severely affect predictive performance, limit representational power, and the ability for kernel 
learning, which is most needed on large datasets \citep{wilson2014thesis}.  New directions for scalable Gaussian processes have involved mini-batches of data 
through stochastic variational inference \citep{hensman2013uai} and distributed learning \citep{deisenroth2015distributed}.
While these approaches are promising, inference can undergo severe approximations, and a small number of inducing points are 
still required. Indeed, stochastic variational approaches scale as $\mathcal{O}(m^3)$.

In this paper, we introduce a new framework for \emph{massively scalable Gaussian processes} (MSGP), which provides
near-exact $\mathcal{O}(n)$ inference and learning and $\mathcal{O}(1)$ test time predictions, and does not require 
distributed learning or severe assumptions.  Our approach builds on 
the recently introduced KISS-GP framework \citep{wilsonnickisch2015}, with several significant advances which enable its use on massive
datasets.  In particular, we provide:
\begin{itemize}
\item Near-exact $\mathcal{O}(1)$ mean and variance predictions.  By contrast, standard GPs and KISS-GP cost
$\mathcal{O}(n)$ for the predictive mean and $\mathcal{O}(n^2)$ for the predictive variance per test point.  Moreover, inducing point and 
finite basis expansions \citep[e.g.,][]{quinonero2005unifying, lazaro2010sparse, yang2015carte} cost $\mathcal{O}(m)$ and $\mathcal{O}(m^2)$ per test point.
\item Circulant approximations which (i) integrate Kronecker and Toeplitz structure, (ii) enable extremely and accurate fast log determinant evaluations for kernel learning, and (iii) increase the speed of Toeplitz methods on problems with 1D predictors.
\item The ability to exploit more general block-Toeplitz-Toeplitz-block (BTTB) structure, which enables fast and exact inference and learning in cases where multidimensional Kronecker structure is not present.
\item Projections which help alleviate the limitation of Kronecker methods to low-dimensional input spaces.
\item Code will be available as part of the GPML package \citep{rasmussen10gpml}, with demonstrations 
at \url{http://www.cs.cmu.edu/~andrewgw/pattern}.
\end{itemize}

We begin by briefly reviewing Gaussian processes, structure exploiting inference, and KISS-GP, in sections \ref{sec: gps} - \ref{sec: kiss}.  We then introduce our MSGP approach in section \ref{sec: msgp}.  We demonstrate the scalability and accuracy of MSGP in the experiments of section \ref{sec: experiments}.  We conclude in section \ref{sec: discussion}.

\section{Gaussian Processes}
\label{sec: gps}

We briefly review Gaussian processes (GPs), and the computational requirements for 
predictions and kernel learning.  \citet{rasmussen06} contains a full treatment of GPs.

We assume a dataset $\mathcal{D}$ of $n$ input (predictor) vectors
$X = [\bm{x}_1,\dots,\bm{x}_n]$, each of dimension $D$,
corresponding to a $n \times 1$ vector of targets 
$\bm{y} = [y(\bm{x}_1),\dots,y(\bm{x}_n)]^{\top}$.  If 
$f(\bm{x}) \sim \mathcal{GP}(\mu,k_{\bm{\theta}})$, then any
collection of function values $\bm{f}$ has a joint Gaussian
distribution,
\begin{align}
 \bm{f} = f(X) = [f(\bm{x}_1),\dots,f(\bm{x}_n)]^{\top} \sim \mathcal{N}(\bm{\mu}_X,K_{X,X}) \,,  \label{eqn: gpdef}
\end{align}
with mean vector and covariance matrix defined by the mean vector and covariance function of
the Gaussian process: $(\bm{\mu}_X)_i = \mu(x_i)$, and $(K_{X,X})_{ij} = k_\bm{\theta}(\bm{x}_i,\bm{x}_j)$.
The covariance function $k_\bm{\theta}$ is parametrized by $\bm{\theta}$.  Assuming additive 
Gaussian noise, $y(\bm{x})|f(\bm{x}) \sim \mathcal{N}(y(\bm{x}); f(\bm{x}),\sigma^2)$,
then the predictive distribution of the GP evaluated at the
$n_*$ test points indexed by $X_*$, is given by
\begin{align}
 \bm{f}_*|X_*,&X,\bm{y},\bm{\theta},\sigma^2 \sim \mathcal{N}(\mathbb{E}[\bm{f}_*],\text{cov}(\bm{f}_*)) \,, \label{eqn: fullpred}  \\  
 \mathbb{E}[\bm{f}_*] &= \bm{\mu}_{X_*}  + K_{X_*,X}[K_{X,X}+\sigma^2 I]^{-1}\bm{y}\,,   \notag \\ 
 \text{cov}(\bm{f}_*) &= K_{X_*,X_*} - K_{X_*,X}[K_{X,X}+\sigma^2 I]^{-1}K_{X,X_*} \,.  \notag
\end{align}
$K_{X_*,X}$ represents the $n_* \times n$ matrix of covariances between 
the GP evaluated at $X_*$ and $X$, and all other covariance matrices 
follow the same notational conventions.  $\bm{\mu}_{X_*}$ is the $n_* \times 1$ mean vector,
and $K_{X,X}$ is the $n \times n$ covariance matrix evaluated at training inputs $X$.
All covariance matrices implicitly depend on the kernel hyperparameters $\bm{\theta}$.

The marginal likelihood of the targets $\bm{y}$  is given by 
\begin{equation}
 \log p(\bm{y} | \bm{\theta}, X) \propto -\frac12[\bm{y}^{\top}(K_{\bm{\theta}}+\sigma^2 I)^{-1}\bm{y} + \log|K_{\bm{\theta}} + \sigma^2 I|]\,,  \label{eqn: mlikeli}
\end{equation}
where we have used $K_{\bm{\theta}}$ as shorthand for $K_{X,X}$ given $\bm{\theta}$.  Kernel learning is performed by optimizing Eq.~\eqref{eqn: mlikeli} with respect to $\bm{\theta}$.

The computational bottleneck for inference is solving the linear system
$(K_{X,X}+\sigma^2 I)^{-1}\bm{y}$, and for kernel learning is computing
the log determinant $\log|K_{X,X}+ \sigma^2 I|$.  Standard
procedure is to compute the Cholesky decomposition of the 
$n \times n$ matrix $K_{X,X}$, which
requires $\mathcal{O}(n^3)$ operations and $\mathcal{O}(n^2)$ storage.
Afterwards, the predictive mean and
variance of the GP cost respectively $\mathcal{O}(n)$ and $\mathcal{O}(n^2)$
per test point $\bm{x}_*$.

\section{Structure Exploiting Inference}
\label{sec: structure}

Structure exploiting approaches make use of existing structure in $K_{X,X}$ to accelerate 
inference and learning.  These approaches benefit from often exact predictive accuracy,
and impressive scalability, but are inapplicable to most problems due to severe grid 
restrictions on the data inputs $X$.  We briefly review Kronecker and Toeplitz structure.

\subsection{Kronecker Structure}
\label{sec: kronecker}

Kronecker (tensor product) structure arises when we have multidimensional inputs i.e. $P>1$ on a rectilinear grid, 
$\bm{x} \in \mathcal{X}_1 \times \dots \times \mathcal{X}_P$, and a product kernel across
dimensions $k(\bm{x}_i,\bm{x}_j) = \prod_{p=1}^{P} k(\bm{x}_i^{(p)},\bm{x}_j^{(p)})$. 
In this case, $K = K_1 \otimes \dots \otimes K_P$.  One can then compute the eigendecomposition
of $K = QVQ^{\top}$ by separately taking the eigendecompositions of the much smaller 
$K_1, \dots, K_P$.  Inference and learning then proceed via 
$(K+\sigma^2 I)^{-1}\bm{y} = (QVQ^{\top} + \sigma^2 I)^{-1}\bm{y} = Q(V+\sigma^2 I)^{-1} Q^{\top} \bm{y}$,
and $\log |K + \sigma^2 I | = \sum_i \log(V_{ii} + \sigma^2)$, where $Q$ is an orthogonal matrix 
of eigenvectors, which also decomposes a Kronecker product (allowing for fast MVMs), and $V$ is 
a diagonal matrix of eigenvalues and thus simple to invert.   Overall, for $m$ grid data points, and
$P$ grid dimensions, inference and learning cost 
$\mathcal{O}(Pm^{1+\frac{1}{P}})$ operations (for $P>1$) and $\mathcal{O}(Pm^{\frac{2}{P}})$ storage 
\citep{saatchi11, wilsonkernel2014}. Unfortunately, there is no efficiency gain for 1D inputs (e.g., time series).

\subsection{Toeplitz Structure}
\label{sec: toeplitz}

A covariance matrix constructed from a stationary kernel $k(x,z) = k(x-z)$ on a
1D regularly spaced grid has Toeplitz structure.  Toeplitz matrices $T$ have constant diagonals, 
$T_{i,j} = T_{i+1,j+1}$.  Toeplitz structure has been exploited for GP inference
\citep[e.g.,][]{zhang05toepGP,cunningham2008fast} in $\mathcal{O}(n \log n)$ computations.
Computing $\log |T|$, and the predictive variance for a single test point,
requires $\mathcal{O}(n^2)$ operations (although finite support can be exploited \citep{storkey99truncCov})
thus limiting Toeplitz methods to about $n < 10,000$ points when kernel learning is required.
Since Toeplitz methods are limited to problems with 1D inputs (e.g., time series), they complement 
Kronecker methods, which exploit multidimensional grid structure.

\section{KISS-GP}
\label{sec: kiss}

% derivation from new perspective
Recently, \citet{wilsonnickisch2015} introduced a fast Gaussian process method
called \emph{KISS-GP}, which performs \emph{local} kernel interpolation, in combination
with inducing point approximations \citep{candela2005} and structure exploiting 
algebra \citep[e.g.,][]{saatchi11, wilson2014thesis}.  

Given a set of $m$ inducing points $U = [\mathbf{u}_i]_{i=1 \dots m}$, 
\citet{wilsonnickisch2015} propose to approximate the $n \times m$ matrix
$K_{X,U}$ of cross-covariances between the training inputs $X$ and inducing
inputs $U$ as $\tilde{K}_{X,U} = W_{X} K_{U,U}$, where $W_{X}$ is an
$n \times m$ matrix of interpolation weights.  One can then approximate
$K_{X,Z}$ for any points $Z$ as ${K}_{X,Z} \approx \tilde{K}_{X,U} W_{Z}^{\top}$.
Given a user-specified kernel $k$, this \emph{structured kernel interpolation} (SKI) 
procedure \citep{wilsonnickisch2015} gives rise to the fast approximate kernel 
\begin{align}
k_{\text{SKI}}(\bm{x},\bm{z}) = \bm{w}_{\bm{x}} K_{U,U} \bm{w}_{\bm{z}} ^{\top} \,,
\end{align}
for any single inputs $\bm{x}$ and $\bm{z}$.  The $n \times n$ training covariance matrix
$K_{X,X}$ thus has the approximation
\begin{align}
K_{X,X} \approx W_{X} K_{U,U} W_{X}^{\top} = K_{\text{SKI}} =: \tilde{K}_{X,X} \,. \label{eqn: ski}
\end{align}
\citet{wilsonnickisch2015} propose to 
perform \emph{local} kernel interpolation, in a method called KISS-GP, 
for extremely sparse 
interpolation matrices.  For example, if we are performing local cubic
\citep{keys1981} interpolation for $d$-dimensional input data,
$W_X$ and $W_Z$ contain only $4^d$ non-zero entries per row.

Furthermore, \citet{wilsonnickisch2015} show that classical inducing point methods
can be re-derived within their SKI framework as global interpolation with a noise 
free GP and non-sparse interpolation weights.  For example, the subset of 
regression (SoR) inducing point method effectively uses the kernel 
${k}_{\text{SoR}} (\bm{x}, \bm{z}) = K_{\bm{x},U} K_{U,U}^{- 1} K_{U,\bm{z}}$
\citep{candela2005}, and thus has interpolation weights 
$\bm{w}_{\text{SoR} (\bm{x})} = K_{\bm{x},U} K_{U,U}^{-1}$ within the SKI framework.

GP inference and learning can be performed in $\mathcal{O}(n)$
using KISS-GP, a significant advance over the more standard $\mathcal{O}(m^2 n)$ 
scaling of fast GP methods \citep{quinonero2005unifying, lazaro2010sparse}.
Moreover, \citet{wilsonnickisch2015} show how -- when performing \emph{local}
kernel interpolation -- one can achieve close to linear scaling 
with the number of inducing points $m$ by placing these points $U$ on a rectilinear grid, and then exploiting 
Toeplitz or Kronecker structure in $K_{U,U}$ \citep[see, e.g.,][]{wilson2014thesis}, without requiring that the data inputs 
$X$ are on a grid.  Such scaling with $m$ compares favourably to the $\mathcal{O}(m^3)$ operations for stochastic variational approaches 
\citep{hensman2013uai}.  Allowing for large $m$ enables near-exact performance, and large scale kernel learning.

In particular, for inference we can solve $(K_{\text{SKI}} + \sigma^2 I)^{-1}\bm{y}$, by performing linear
conjugate gradients, an iterative procedure which depends only on matrix vector multiplications (MVMs) 
with $(K_{\text{SKI}} + \sigma^2 I)$.  Only $j \ll n$ iterations are required for convergence up to 
machine precision, and the value of $j$ in practice depends on the conditioning of $K_{\text{SKI}}$ 
rather than $n$.  MVMs with sparse $W$ (corresponding to local interpolation) cost $\mathcal{O}(n)$, 
and MVMs exploiting structure in $K_{U,U}$ are roughly linear in $m$.  Moreover, we can efficiently
approximate the eigenvalues of $K_{\text{SKI}}$ to evaluate $\log|K_{\text{SKI}}+\sigma^2 I|$, for kernel learning, 
by using fast structure exploiting eigendecompositions of $K_{U,U}$.  Further details are in 
\citet{wilsonnickisch2015}.

\section{Massively Scalable Gaussian Processes}
\label{sec: msgp}

We introduce massively scalable Gaussian processes (MSGP), which 
significantly extend KISS-GP, inducing point, and structure exploiting 
approaches, for:
(1) $\mathcal{O}(1)$ test predictions (section \ref{sec: fast}); 
(2) circulant log determinant approximations which (i) unify Toeplitz and Kronecker structure;
    (ii) enable extremely fast marginal likelihood evaluations (section \ref{sec: circulant}); 
    and (iii) extend KISS-GP and Toeplitz methods for scalable kernel learning in D=1 input dimensions,
    where one cannot exploit multidimensional Kronecker structure for scalability.
(3) more general BTTB structure, which enables fast exact multidimensional inference without 
requiring Kronecker (tensor) decompositions (section \ref{sec: BCCB}); and, 
(4) projections which enable KISS-GP to be used with structure exploiting
approaches for $D \gg 5$ input dimensions, and increase the expressive
power of covariance functions (section \ref{sec: project}).

\subsection{Fast Test Predictions}
\label{sec: fast}

While \citet{wilsonnickisch2015} propose fast 
$\mathcal{O}(n)$ inference and learning, test time predictions are 
the same as for a standard GP -- namely, $\mathcal{O}(n)$ for the 
predictive mean and $\mathcal{O}(n^2)$ for the predictive variance per single test point 
$\bm{x}_*$.  Here we show how to obtain $\mathcal{O}(1)$ test time
predictions by efficiently approximating latent mean and variance 
of $\bm{f}_*$. 
For a Gaussian likelihood, the predictive distribution
for $\bm{y}_*$ is given by the relations
$\mathbb{E}[\bm{y}_*]=\mathbb{E}[\bm{f}_*]$ and
$\text{cov}(\bm{y}_*)=\text{cov}(\bm{f}_*)+\sigma^2 I$.

We note that the methodology here for fast test predictions does not rely on having
performed inference and learning in any particular way: it can be applied to \emph{any}
trained Gaussian process model, including a full GP, inducing points methods such as
FITC \citep{snelson2006sparse}, or the \emph{Big Data GP} \citep{hensman2013uai}, or 
finite basis models \citep[e.g.,][]{yang2015carte, lazaro2010sparse, Rahimi07,le2013fastfood, williams2001}.

\subsubsection{Predictive Mean}

Using structured kernel interpolation on $K_{X,X}$, we 
approximate the predictive mean 
$\mathbb{E}[\bm{f}_*]$ of Eq.~\eqref{eqn: fullpred} for a
set of $n_*$ test inputs $X_*$ as 
$\mathbb{E}[\bm{f}_*] \approx \bm{\mu}_{X_*}  + K_{X_*,X} \bm{\tilde{\alpha}}$, 
where $\bm{\tilde{\alpha}} = [\tilde{K}_{X,X} + \sigma^2 I]^{-1} \bm{y}$
is computed as part of training using linear conjugate gradients (LCG).
We propose to successively apply structured kernel interpolation 
on $K_{X_*,X}$ for 
\begin{align}
\mathbb{E}[\bm{f}_*] \approx \mathbb{E}[\bm{\tilde{f}}_*] &= \bm{\mu}_{X_*}  +  \tilde{K}_{X_*,X}  \bm{\tilde{\alpha}}\,, \\
&= \bm{\mu}_{X_*}  + W_* K_{U,U} W^{\top}  \bm{\tilde{\alpha}} \,, 
\end{align}
where
$W_*$ and $W$ are respectively $n_* \times m$ and $n \times m$
sparse interpolation matrices, containing only $4$ non-zero 
entries per row if performing local cubic interpolation (which we 
henceforth assume).
The term $K_{U,U} W^{\top} \bm{\tilde{\alpha}}$
is pre-computed during training, 
taking only two MVMs in addition to the LCG computation
required to obtain $\bm{\tilde{\alpha}}$.\footnote{We exploit the structure of $K_{U,U}$ 
for extremely efficient MVMs, which we will discuss in detail in section 
\ref{sec: circulant}.}
Thus the only computation at test time is multiplication
with sparse $W_*$, which costs $\mathcal{O}(n_*)$ operations, leading to 
$\mathcal{O}(1)$ operations per test point $\bm{x}_*$.

\subsubsection{Predictive Variance}
\label{sec: predvar}

Practical applications typically do not require the full predictive covariance
matrix $\text{cov}(\bm{f}_*)$ of Eq.~\eqref{eqn: fullpred} for a set of $n_*$
test inputs $X_*$, but rather focus on the predictive variance,
\begin{align}
  \bm{v}_*=\text{diag}[\text{cov}(\bm{f}_*)]=\text{diag}(K_{X_*,X_*}) - \bm{\nu}_*,
\end{align}
where $\bm{\nu}_*=\text{diag}( K_{X_*,X}[K_{X,X}+\sigma^2 I]^{-1}K_{X,X_*} )$,
the explained variance, is approximated by local interpolation from the explained
variance on the grid $U$
\begin{align}
  \bm{\nu}_* \approx W_* \bm{\tilde{\nu}}_U, \: \bm{\tilde{\nu}}_U=\text{diag}(\tilde{K}_{U,X}A^{-1}\tilde{K}_{X,U}).
\end{align}
using the interpolated covariance $A=\tilde{K}_{X,X}+\sigma^2 I$.
Similar to the predictive mean -- once $\bm{\tilde{\nu}}_U$ is precomputed -- 
we only require a multiplication with the sparse interpolation weight matrix $W_*$,
leading to $\mathcal{O}(1)$ operations per test point $\bm{x}_*$.

Every $[\bm{\tilde{\nu}}_U]_i$ requires the solution to a linear system of size $n$, which is 
computationally taxing. To \emph{efficiently} precompute $\bm{\tilde{\nu}}_U$, we instead employ a stochastic estimator 
$\bm{\hat{\nu}}_U$ \citep{papandreou11diaginv}, based on the observation that 
$\bm{\tilde{\nu}}_U$ is the variance of the projection 
$\tilde{K}_{U,X}\bm{r}$ of the Gaussian random variable $\bm{r}\sim\mathcal{N}(\bm{0},A^{-1})$.
We draw $n_s$ Gaussian samples $\bm{g}^m_i\sim\mathcal{N}(\bm{0},I)$, $\bm{g}^n_i\sim\mathcal{N}(\bm{0},I)$ and solve $A \bm{r}_i= W V \sqrt{E} V^{\top} \bm{g}^m_i + \sigma\bm{g}^n_i$ with LCG, where $K_{U,U} = V E V^{\top}$ is the eigendecomposition of the covariance evaluated on the grid, which can be 
computed efficiently by exploiting Kronecker and Toeplitz structure (sections \ref{sec: structure} and \ref{sec: circulant}).

 The overall (unbiased) estimate \citep{papandreou11diaginv} is obtained by clipping
\begin{align}
  \bm{v}_*\approx\bm{\hat{v}}_* = \max [ \bm{0}, 
                            \bm{k}_* - W_* \sum_{i=1}^{n_s} (\tilde{K}_{U,X} \bm{r}_i)^2 ],
\end{align}
where the square is taken element-wise. \citet{papandreou11diaginv} suggest to use a value of
$n_s=20$ Gaussian samples $\bm{r}_i$ which corresponds to a relative error 
$||\bm{\hat{\nu}}_U-\bm{\tilde{\nu}}_U||/||\bm{\tilde{\nu}}_U||$ of $0.36$.

\subsection{Circulant Approximation}
\label{sec: circulant}

Kronecker and Toeplitz methods (section \ref{sec: structure}) are greatly 
restricted by requiring that the data inputs $X$ are located on a grid.  We lift this
restriction by creating structure in $K_{U,U}$, with the unobserved inducing
variables $U$, as part of the structured kernel interpolation framework 
described in section \ref{sec: kiss}.  

Toeplitz methods apply only to 1D problems, and Kronecker methods require multidimensional 
structure for efficiency gains.  Here we present a circulant approximation 
to \emph{unify} the \emph{complementary} benefits of Kronecker and Toeplitz methods, and to greatly 
scale marginal likelihood evaluations of Toeplitz based methods, while not requiring
any grid structure in the data inputs $X$.

If $U$ is a \emph{regularly spaced} multidimensional grid, and we use a stationary
product kernel (e.g., the RBF kernel), then $K_{U,U}$ decomposes as a Kronecker 
product of \emph{Toeplitz} (section \ref{sec: toeplitz}) matrices:
\begin{align}
K_{U,U} = T_1 \otimes \dots \otimes T_P \,.  \label{eqn: krontoep}
\end{align}
Because the algorithms which leverage Kronecker structure in Gaussian processes
require eigendecompositions of the constituent matrices, computations involving the 
structure in Eq.~\eqref{eqn: krontoep} are no faster than if the Kronecker product were
over arbitrary positive definite matrices: this nested Toeplitz structure, which is often present in 
Kronecker decompositions, is wasted.  Indeed, 
while it is possible to efficiently solve linear systems with Toeplitz matrices, there is no 
particularly efficient way to obtain a full eigendecomposition.

Fast operations with an $m \times m$ Toeplitz matrix $T$ can be obtained through its
relationship with an $a \times a$ \emph{circulant matrix}.  Symmetric circulant matrices $C$ 
are Toeplitz matrices where the first column $\bm{c}$ is given by a circulant vector:
$\bm{c} = [c_{1},c_{2},c_{3},..,c_{3},c_{2}]^{\top}$.  
Each subsequent column is shifted one position from the next.  In other words, 
$C_{i,j} = \bm{c}_{| j - i | \text{ mod } a}$.  Circulant matrices are computationally 
attractive because their eigendecomposition is given by
\begin{align}
C = F^{-1} \text{diag}(F\bm{c}) F \,,  \label{eqn: circeig}
\end{align}
where $F$ is the discrete Fourier transform (DFT): $F_{jk} = \exp(-2 j k \pi i / a)$.
The eigenvalues of $C$ are thus given by the DFT of its first column, and the eigenvectors are 
proportional to the DFT itself (the $a$ roots of unity).  The log determinant of $C$ -- the sum of log 
eigenvalues --  can therefore be computed from a single fast Fourier transform (FFT) which costs
$\mathcal{O}(a \log a)$ operations and $\mathcal{O}(a)$ memory.  Fast matrix vector products can
be computed at the same asymptotic cost through Eq.~\eqref{eqn: circeig}, requiring two FFTs
(one FFT if we pre-compute $F\bm{c}$), one inverse FFT, and one inner product.

In a Gaussian process context, fast MVMs with Toeplitz matrices are typically achieved
through embedding a $m \times m$ Toeplitz matrix $K$ into a $(2m-1) \times (2m-1)$ 
circulant matrix $C$ \citep[e.g.,][]{zhang05toepGP,cunningham2008fast}, with first column
$\bm{c}=[k_{1},k_{2},..,k_{m-1},k_{m},k_{m-1},..,k_{2}]$.   Therefore 
$K = C_{i=1 \dots m,j=1 \dots m}$, and using zero padding and truncation, 
$K \bm{y} = [C\bm{y}, \bm{0}]^{\top}_{i=1 \dots m}$, where the circulant MVM $C \bm{y}$ can be
computed efficiently through FFTs.  GP inference can then be achieved through 
LCG, solving $K^{-1} \bm{y}$ in an iterative procedure which only involves MVMs,
and has an asymptotic cost of $\mathcal{O}(m \log m)$ computations.  
The log determinant and a single predictive variance, however, require $\mathcal{O}(m^2)$
computations.

To speed up LCG for solving linear Toeplitz systems, one can use circulant pre-conditioners
which act as approximate inverses.  One wishes to minimise the distance between the 
preconditioner $C$ and the Toeplitz matrix $K$, $\text{arg min}_C d(C,K)$.
Three classical pre-conditioners include 
$C_{\text{Strang}}=\arg\min_{C}\left\Vert C-K\right\Vert _{1}$
\citep{strang86circ-precond}, 
$C_{\text{T. Chan}}=\arg\min_{C}\left\Vert C-K\right\Vert _{F}$
\citep{chan88circ-precond}, and 
$C_{\text{Tyrtyshnikov}}=\arg\min_{C}\left\Vert I-C^{-1}K\right\Vert _{F}$
\citep{tyrtyshnikov92circ-precond}. 

A distinct line of research was explored 
more than 60 years ago in the context of statistical inference 
over spatial
processes \citep{whittle54planeStatProcess}. The circulant Whittle approximation
$\text{circ}_{\text{Whittle}}(\bm{k})$ is given by truncating the sum
\[
\left[\text{circ}_{\text{Whittle}}(\bm{k})\right]_i=\sum_{j\in\mathbb{Z}}k_{i+jm}\text{ or }c(t)=\sum_{j\in\mathbb{Z}}k(t+jm\Delta u)
\]
i.e. we retain $\sum_{j=-w}^{w}k_{i+jm}$ only.  

Positive definiteness of $C=\text{toep}(\bm{c})$ for the complete sum is guaranteed by 
construction \citep[Section 2]{guinness14circEmbedInference}.
For large lattices, the approach is often used due to its accuracy
and favourable asymptotic properties such as consistency, efficiency
and asymptotic normality \citep{lieberman09asymptotictheory}. In fact,
the circulant approximation $c_{i}$ is asymptotically equivalent
to the initial covariance $k_{i}$, see \citet[Lemma 4.5]{gray05toepcirc},
hence the logdet approximation inevitably converges to the exact value.

We use 
\[
\log|\text{toep}(\bm{k})+\sigma^{2}I|\approx\bm{1}^{\top}\log(F\bm{c}+\sigma^{2}\bm{1}),
\]
where we threshold $\bm{c}=F^{\textsf{H}}\max(F\text{circ}(\bm{k}),\bm{0})$.
$\bm{1}$ denotes a vector $\bm{1} = [1,1,\dots,1]^{\top}$, $F^{H}$ is the conjugate transpose
of the Fourier transform matrix.   $\text{circ}(\bm{k})$
denotes one of the T. Chan, Tyrtyshnikov,
Strang, Helgason or Whittle approximations. 

\begin{figure}
\begin{centering}
\includegraphics[bb=45bp 15bp 880bp 590bp,clip,width=1\columnwidth]{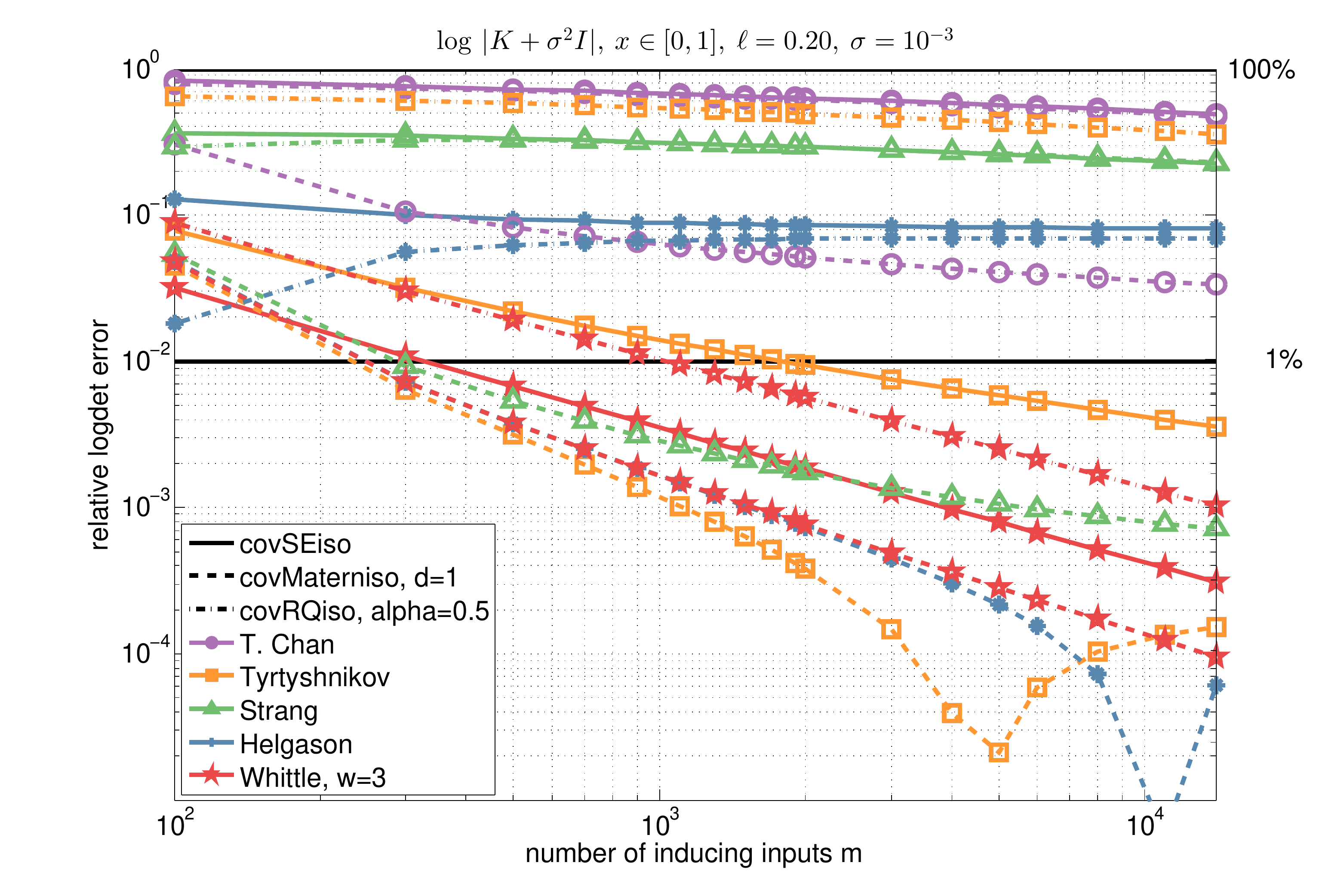}
\par\end{centering}
\protect\caption{\label{fig:circ_benchmark}Benchmark of different circulant approximations illustrating consistent good quality of the Whittle embedding.
covSE, covMatern, and covRQ are as defined in \citet{rasmussen06}.}
\end{figure}

The particular circulant approximation pursued can have a dramatic practical effect
on performance.  In an empirical comparison (partly shown
in Figure \ref{fig:circ_benchmark}), we verified that the Whittle
approximations yields consistently accurate approximation results
over several covariance functions $k(x-z)$, lengthscales $\ell$
and noise variances $\sigma^{2}$ decaying with grid size $m$ and 
below $1\%$ relative error for $m>1000$.

\subsection{BCCB Approximation for BTTB}
\label{sec: BCCB}
There is a natural extension of the circulant approximation of section \ref{sec: circulant}
to multivariate $D>1$ data. A translation invariant covariance function
$k(\bm{x},\bm{z})=k(\bm{x}-\bm{z})$ evaluated at data points $\bm{u}_i$ organised
on a regular grid of size  $n_1 \times n_2 \times .. \times n_D$ result in a symmetric
block-Toeplitz matrix with Toeplitz blocks (BTTB), which generalises Toeplitz matrices.
Unlike with Kronecker methods, the factorisation of a covariance function is not required for
this structure.
Using a dimension-wise circulant embedding of size
$(2n_1-1) \times (2n_2-1) \times .. \times (2n_D-1)$, fast MVMs can be accomplished using
the multi-dimensional Fourier transformation $F=F_1 \otimes F_2 \otimes .. \otimes F_D$ by
applying Fourier transformations $F_d$ along each dimension rendering fast inference using 
LCG feasible.
Similarly, the Whittle approximation to the log-determinant can be generalised, where 
the truncated sum for $\text{circ}_{\text{Whittle}}(\bm{k})$ is over $(2w+1)^D$ terms
instead of $2w+1$. As a result, the Whittle approximation $C_{U,U}$ to the 
covariance matrix $K_{U,U}$ is block-circulant with circulant blocks (BCCB).
Fortunately, BCCB matrices have an eigendecomposition $C_{U,U}=F^{\textsf{H}} (F\bm{c})  F$,
where $\bm{c}\in \mathbb{R}^n$ is the Whittle approximation to 
$\bm{k}\in \mathbb{R}^n$, $n = n_1 \cdot n_2 \cdot .. \cdot n_D$ and $F=F_1 \otimes F_2 \otimes .. \otimes F_D$ as defined before.
Hence, all computational and approximation benefits from the Toeplitz case carry over to the
BTTB case. As a result, exploiting the BTTB structure allows to efficiently deal
with multivariate data without requiring a factorizing covariance function.

We note that blocks in the BTTB matrix need not be symmetric.  Moreover, symmetric BCCB matrices
-- in contrast to symmetric BTTB matrices -- are fully characterised by their first column.

\subsection{Projections}
\label{sec: project}

Even if we do not exploit structure in $K_{U,U}$, our framework in section \ref{sec: msgp} provides efficiency gains 
over conventional inducing point approaches, particularly for test time predictions.  However, if we are to place $U$ onto a 
multidimensional (Cartesian product) grid so that $K_{U,U}$ has Kronecker structure, then the total number of inducing points 
$m$ (the cardinality of $U$) grows exponentially with the number of grid dimensions, limiting one to about five or fewer grid dimensions 
for practical applications.  However, we need not limit the applicability of our approach to data inputs $X$ with $D \leq 5$ input dimensions, 
even if we wish to exploit Kronecker structure in $K_{U,U}$.  Indeed, many inducing point approaches suffer from the curse of dimensionality, 
and input projections have provided an effective countermeasure \citep[e.g.,][]{snelson07}.

We assume the $D$ dimensional data inputs $\bm{x} \in \mathbb{R}^{D}$, and inducing points which live in a 
lower $d < D$ dimensional space, $\bm{u} \in \mathbb{R}^{d}$, and are related through the mapping 
$\bm{u} = h(\bm{x},\bm{\omega})$, where we wish to learn the parameters of the mapping $\bm{\omega}$ 
in a \emph{supervised} manner, through the Gaussian process marginal likelihood. Such a representation
is highly general: any deep learning architecture $h(\bm{x},\bm{\omega})$, for example, will ordinarily project into a 
hidden layer which lives in a $d < D$ dimensional space. 

We focus on the supervised learning of linear projections, $P \bm{x} = \bm{u}$, where $P \in \mathbb{R}^{d \times D}$.
Our covariance functions effectively become $k(\bm{x}_i, \bm{x}_j) \to k( P \bm{x}_i, P \bm{x}_j)$, 
$k(\bm{x}_i, \bm{u}_j) \to k(P\bm{x}_i, \bm{u}_j)$, $k(\bm{u}_i,\bm{u}_j) \to k(\bm{u}_i,\bm{u}_j)$.   Starting from the
RBF kernel, for example, 
\begin{align}
k_{\text{RBF}}(Px_i, Px_j) &= \exp \left[ -0.5 (P \bm{x}_i - P\bm{x}_j)^{\top} (P\bm{x}_i - P\bm{x}_j) \right]  \notag \\
&= \exp \left[ -0.5 (\bm{x}_i - \bm{x}_j)^{\top} PP^{\top}  (\bm{x}_i-\bm{x}_j) \right] \,. \notag
\end{align}
The resulting kernel generalises the RBF and ARD kernels, which respectively have spherical and diagonal 
covariance matrices, with a full covariance matrix $\Sigma = PP^{\top}$, which allows for
richer models \citep{vivarelli1998discovering}.  But in our context, this added flexibility has special meaning.
Kronecker methods typically require a kernel which separates as a product across input dimensions 
(section \ref{sec: kronecker}).  Here, we can capture sophisticated correlations between the different dimensions
of the data inputs $\bm{x}$ through the projection matrix $P$, while preserving Kronecker structure in $K_{U,U}$.
Moreover, learning $P$ in a \emph{supervised} manner, e.g., through the Gaussian porcess marginal likelihood, 
has immediate advantages over unsupervised dimensionality
reduction of $\bm{x}$; for example, if only a subset of the data inputs were used in producing the target values, this
structure would not be detected by an unsupervised method such as PCA, but can be learned through $P$.  
Thus in addition to the critical benefit of allowing for applications with $D > 5$ dimensional data inputs, $P$ can 
also enrich the expressive power of our MSGP model.

The entries of $P$ become hyperparameters of the Gaussian process marginal likelihood of Eq.~\eqref{eqn: mlikeli}, and
can be treated in exactly the same way as standard kernel hyperparameters such as length-scale.  One can learn these
parameters through marginal likelihood optimisation.

Computing the derivatives of the log marginal likelihood with respect to the projection matrix requires some care 
under the structured kernel interpolation approximation to the covariance matrix $K_{X,X}$.  We provide the
mathematical details in the appendix \ref{sec: appendix}.

For practical reasons, one may wish to restrict $P$ to be orthonormal or have unit scaling.  We discuss this further 
in section \ref{sec: projexp}.

\section{Experiments}
\label{sec: experiments}

In these preliminary experiments, we stress test MSGP in terms of training and prediction runtime, as well as accuracy,
empirically verifying its scalability and predictive performance.  We also demonstrate the consistency of the model in 
being able to learn supervised projections, for higher dimensional input spaces.  

We compare to exact Gaussian processes, FITC \citep{snelson2006sparse}, Sparse Spectrum Gaussian Processes (SSGP) 
\citep{lazaro2010sparse}, the \emph{Big Data GP} (BDGP) \citep{hensman2013uai}, MSGP with Toeplitz (rather than circulant) methods, and 
MSGP not using the new scalable approach for test predictions (scalable test predictions are described in section \ref{sec: msgp}).

The experiments were executed on a workstation with Intel i7-4770 CPU and 32 GB
RAM. We used a step length of $0.01$ for BDGP based on the values reported by
\citet{hensman2013uai} and a batchsize of $300$. We also evaluated BDGP with a
larger batchsize of $5000$ but found that the results are qualititively
similar. We stopped the stochastic optimization in BDGP when the log-likelihood
did not improve at least by $0.1$ within the last $50$ steps or after $5000$
iterations.

\subsection{Stress tests}

One cannot exploit Kronecker structure in one dimensional inputs for scalability, and Toeplitz methods, which apply to 
1D inputs, are traditionally limited to about $10,000$ points if one requires many marginal likelihood evaluations for kernel 
learning.  Thus to stress test the value of the circulant approximation most transparently, and to give the greatest advantage to 
alternative approaches in terms of scalability with number of inducing points $m$, we initially stress test in a 1D input space, before 
moving onto higher dimensional problems.

In particular, we sample 1D inputs $x$ uniform randomly in $[-10,10]$, so that the data inputs have no grid structure.
We then generate out of class ground truth data $f(x) =  \sin(x) \exp \left(\frac{-x^2}{2 \times 5^2} \right)$ with additive Gaussian 
noise to form $\bm{y}$.  We distribute inducing points on a regularly spaced grid in $[-12,13]$.  

\begin{figure}
\centering{}
\includegraphics[scale=1]{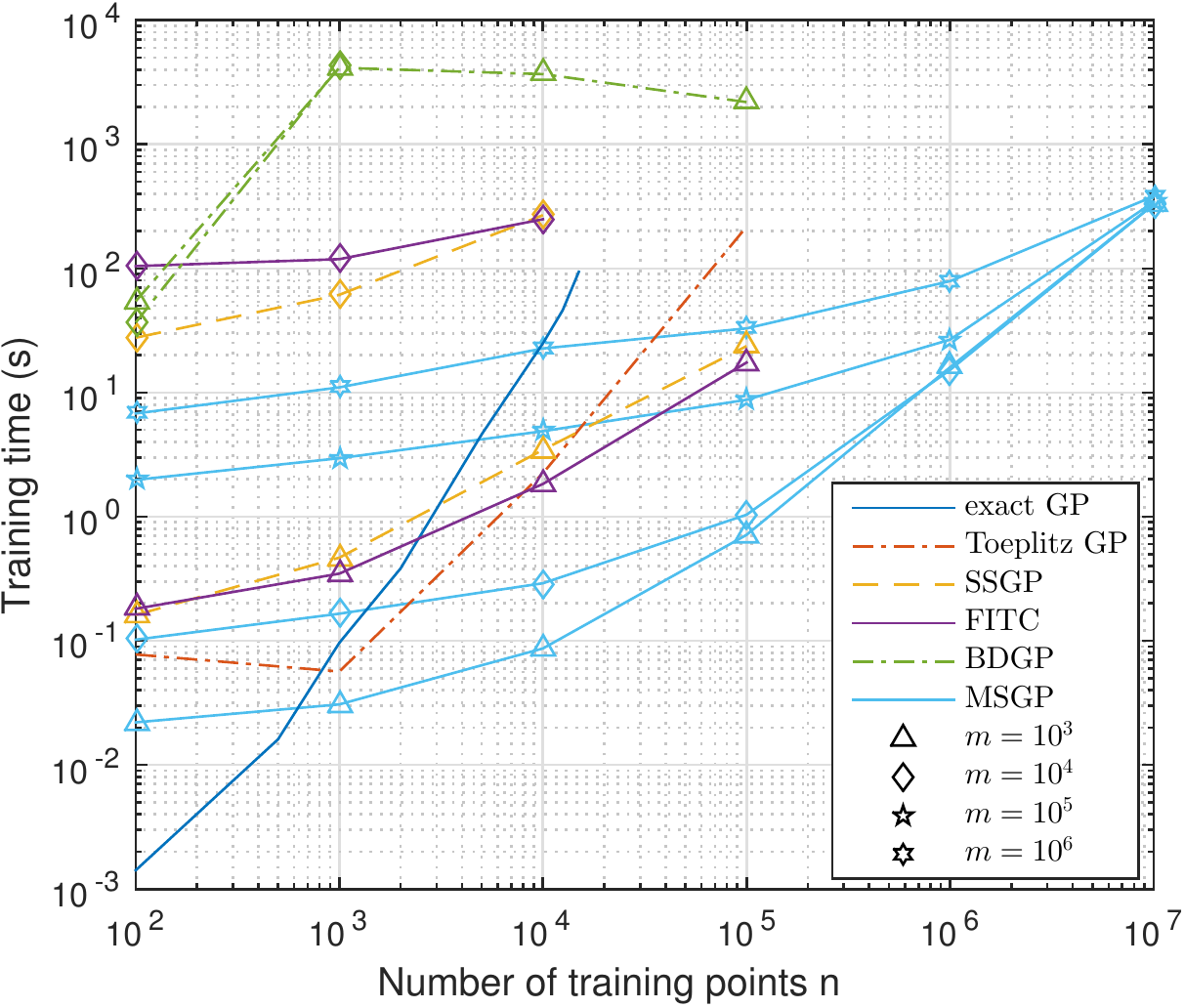}
\protect\caption{Training Runtime Comparison.  We evaluate the runtime, in seconds, to evaluate the marginal likelihood and all relevant derivatives for 
each given method.}
\label{fig: training1k}
\end{figure}

In Figure \ref{fig: training1k} we show the runtime for a marginal likelihood
evaluation as a function of training points $n$, and number inducing points
$m$.
MSGP quickly overtakes the alternatives as $n$ increases past this 
point.  Moreover, the runtimes for MSGP with different numbers of inducing 
points converge quickly with increases in $m$.  By $n=10^7$, MSGP requires
the same training time for $m=10^3$ inducing points as it does for $m=10^6$
inducing points!   Indeed, MSGP is able to 
accommodate an unprecedented number of inducing points, overtaking
the alternatives, which are using $m=10^3$ inducing points, when using
$m=10^6$ inducing points.  Such an exceptionally large number of inducing
points allows for near-exact accuracy, and the ability to retain model structure
necessary for large scale kernel learning.  Note that $m=10^3$ inducing points
(resp. basis functions) is a practical upper limit in alternative approaches (\citet{hensman2013uai},
for example, gives $m \in [50,100]$ as a practical upper bound for conventional inducing
approaches for large $n$). 

We emphasize that although the stochastic optimization of the BDGP can take some
time to converge, the ability to use mini-batches for jointly optimizing the variational parameters 
and the GP hyper parameters, as part of a principled framework, makes BDGP
highly scalable.  Indeed, the methodology in MSGP and BDGP are complementary 
and could be combined for a particularly scalable Gaussian process framework.  

\begin{figure*}
\centering{}
\includegraphics[scale=0.9]{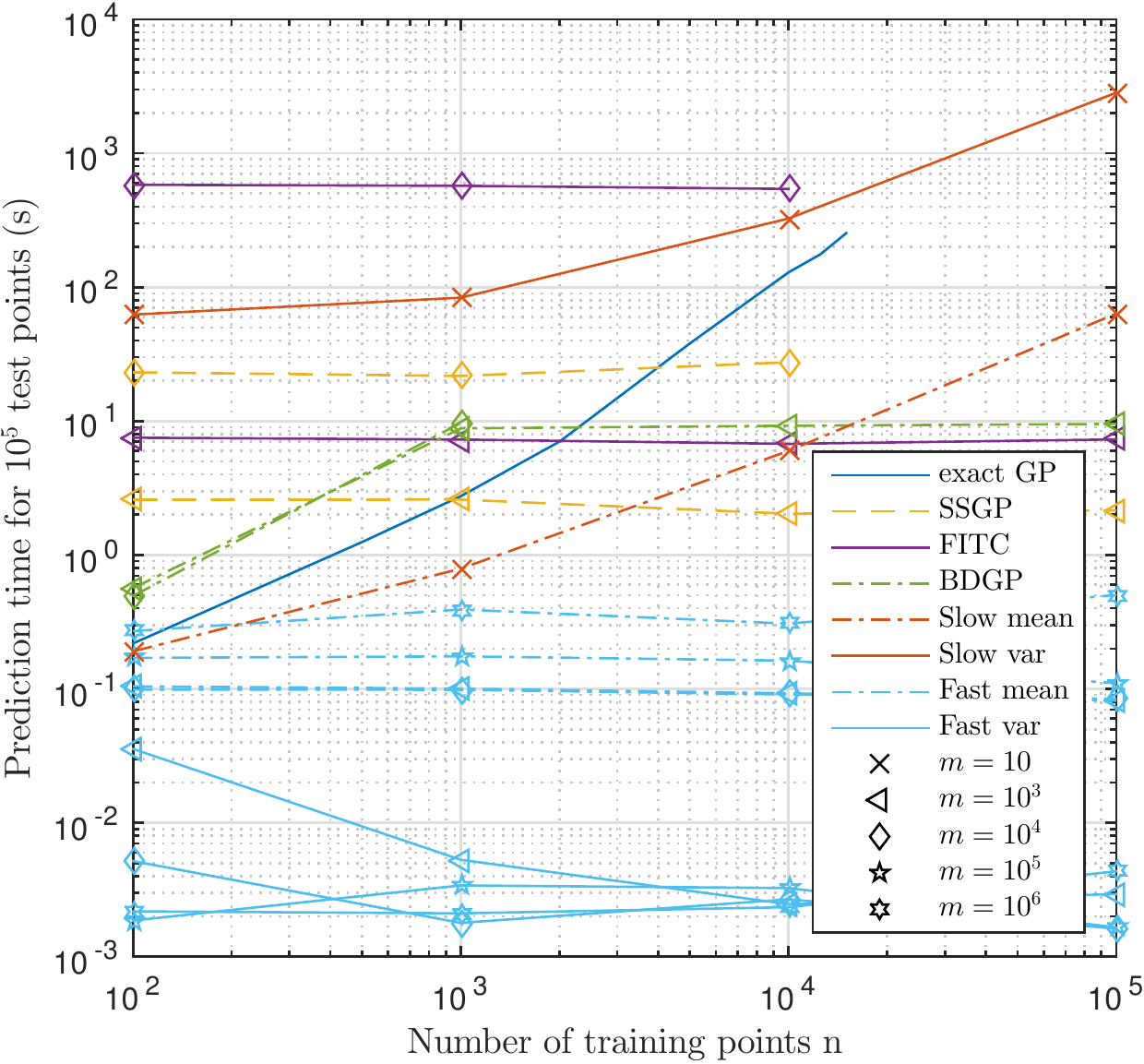} 
\includegraphics[scale=1]{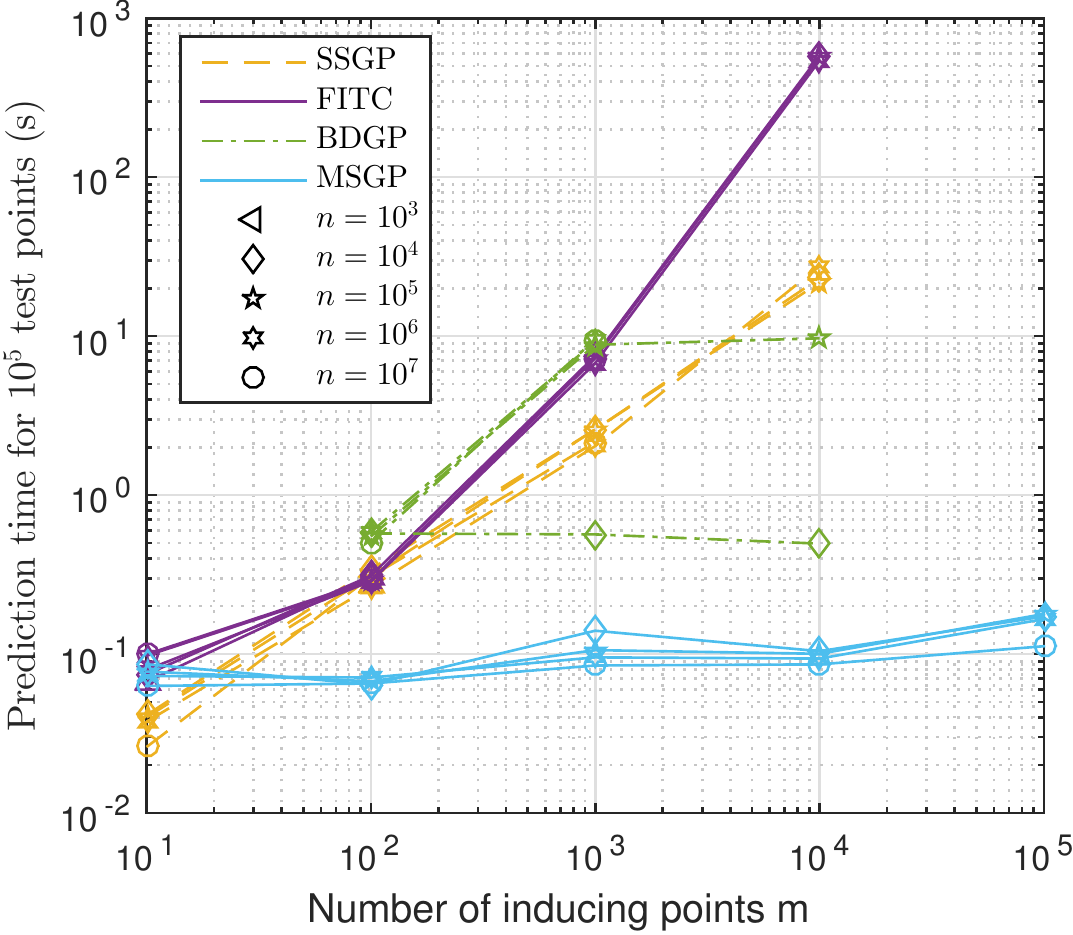}
\protect\caption{Prediction Runtime Comparison.  `slow' mean and var refer to BDGP when using standard Kronecker and Toeplitz
algebra for the predictive mean and variance, without fast local interpolation proposed in section \ref{sec: fast}.}
\label{fig: prediction}
\end{figure*}

In Figure \ref{fig: prediction}, we show that the prediction runtime for MSGP is 
practically independent of both $m$ and $n$, and for any fixed $m,n$ is much faster 
than the alternatives, which depend at least quadratically on $m$ and $n$.  We 
also see that the local interpolation strategy for test predictions, introduced in 
section \ref{sec: fast}, greatly improves upon MSGP using the exact predictive
distributions while exploiting Kronecker and Toeplitz algebra.

\begin{figure}
\centering{}
\includegraphics[scale=.8]{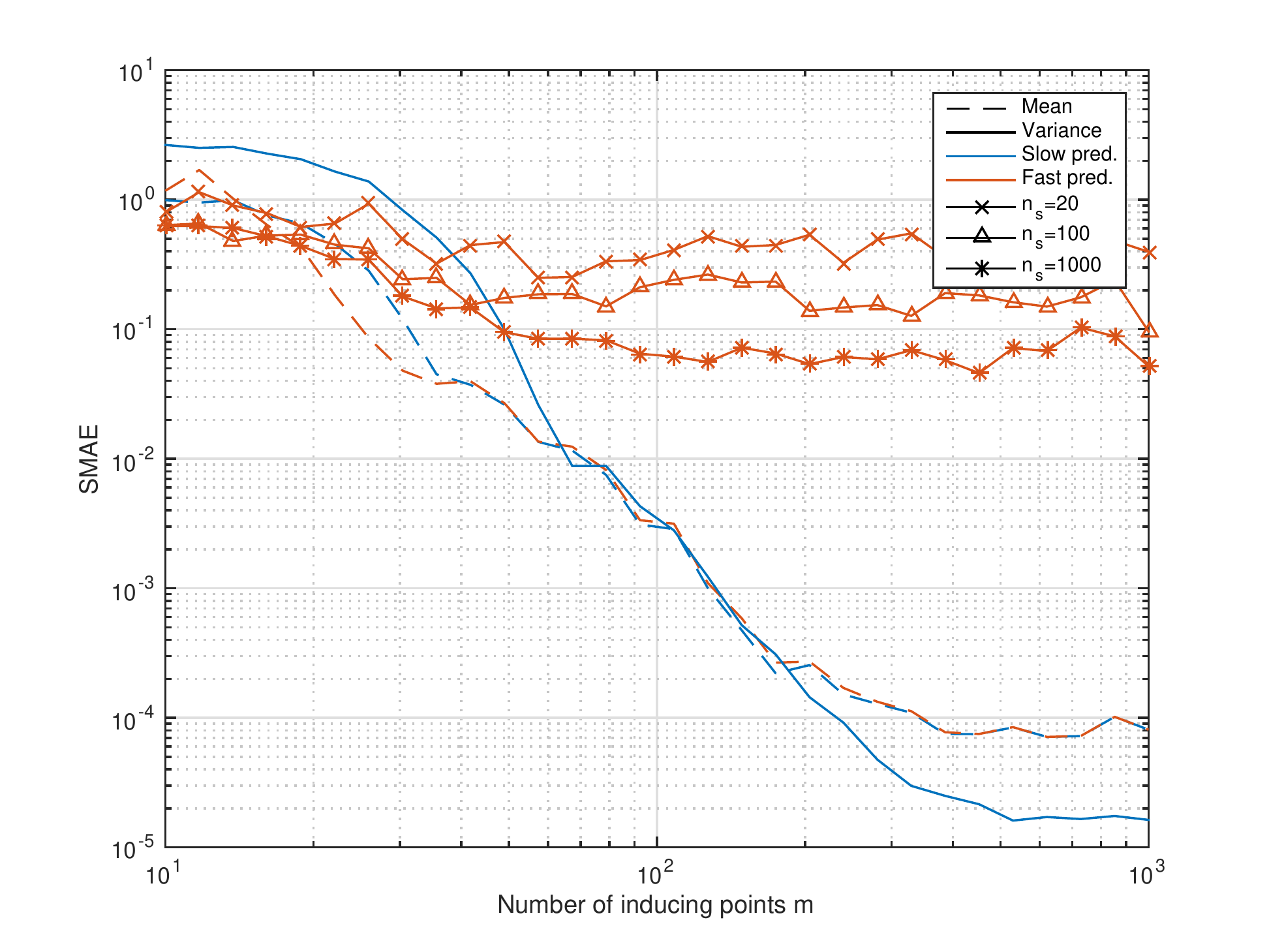}
\protect\caption{Accuracy Comparison. We compare the relative absolute difference of the predictive mean and variance of MSGP both using 
`fast' local kernel interpolation, and the `slow' standard predictions, to exact inference.}
\label{fig: accuracy_tea}
\end{figure}

In Figure \ref{fig: accuracy_tea}, we evaluate the accuracy of the fast mean and variance test predictions in section \ref{sec: fast}, using the same 
data as in the runtime stress tests.  We compare the relative mean absolute error, $\text{SMAE}(\tilde{\bm{y}}_*) = \text{MAE}(\bm{y}_*,\tilde{\bm{y}}_*) / \text{MAE}(\bar{\bm{y}}_*,\bm{y}_*)$, where $\bm{y}_*$ are the true test targets, and $\bar{\bm{y}}_*$ is the mean of the true test targets, for the test predictions $\tilde{\bm{y}}_*$ made by
each method.  We compare the predictive mean and variance of the fast predictions with MSGP using the standard (`slow') predictive equations and Kronecker algebra,
and the predictions made using an exact Gaussian process.  

As we increase the number of inducing points $m$ the quality of predictions are improved, as we expect.  The fast predictive variances, based on local kernel 
interpolation, are not as sensitive to the number of inducing points as the alternatives, but nonetheless have reasonable performance.  We also see that the 
the fast predictive variances are improved by having an increasing number of samples $n_s$ in the stochastic estimator of section \ref{sec: predvar}, which is
most noticeable for larger values numbers of inducing points $m$.  Notably, the fast predictive mean, based on local interpolation, is essentially indistinguishable
from the predictive mean (`slow') using the standard GP predictive equations without interpolation.  Overall, the error of the fast predictions is much less than 
the average variability in the data.

\subsection{Projections}
\label{sec: projexp}

\begin{figure*}
\centering{}
\subfigure[]{\label{fig: proja} \includegraphics[scale=1]{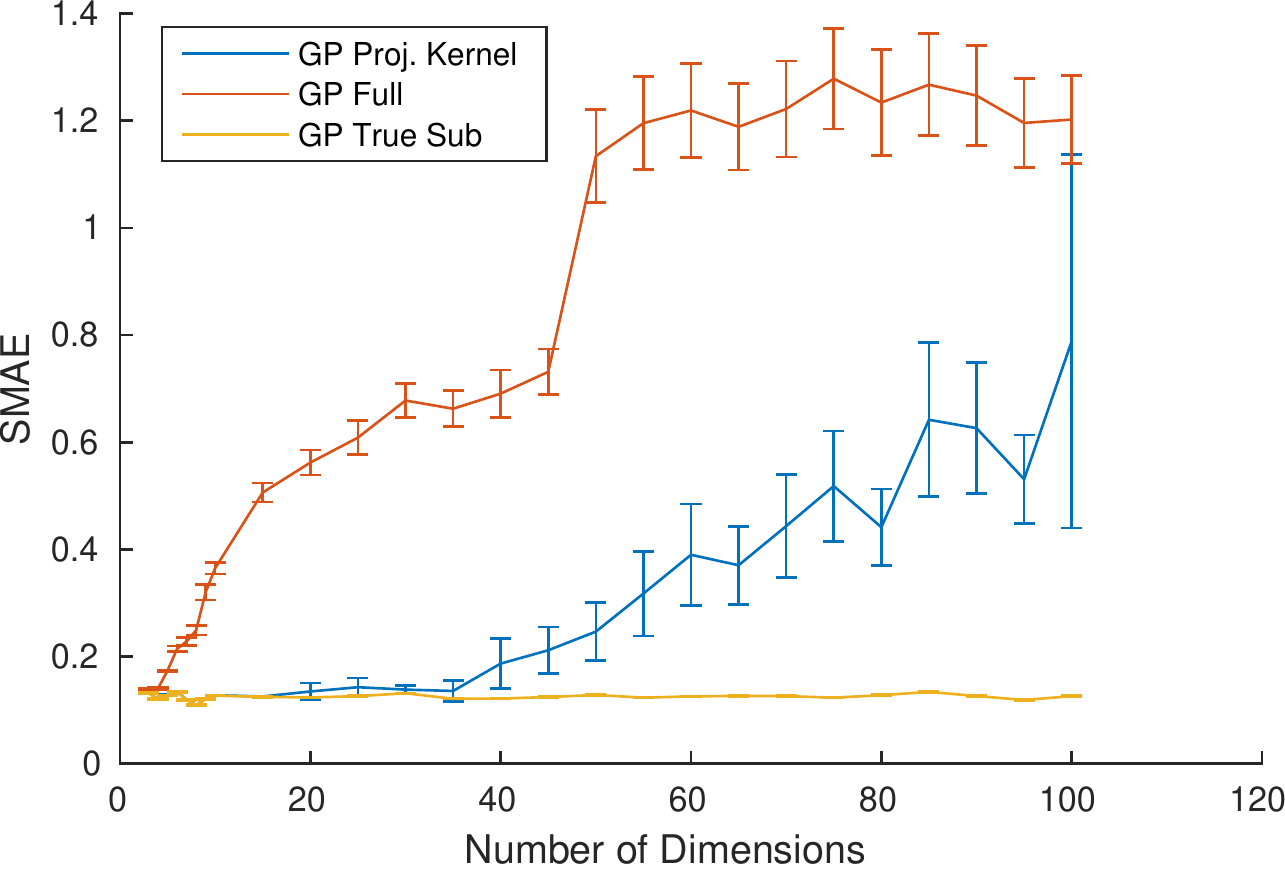}} 
\subfigure[]{\label{fig: projb} \includegraphics[scale=1]{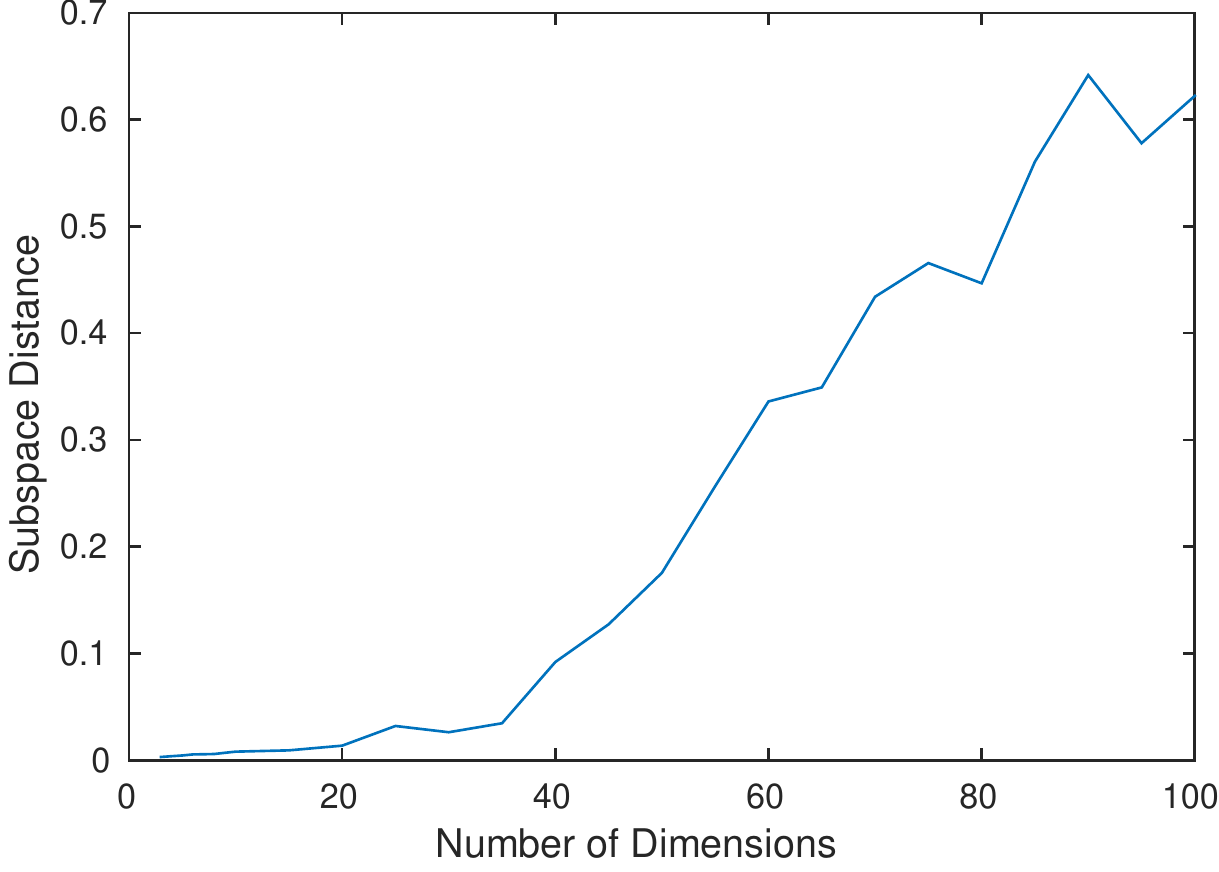}}
\protect\caption{Synthetic Projection Experiments}
\end{figure*}

Here we test the consistency of our approach in section \ref{sec: project} for recovering ground truth projections,
and providing accurate predictions, on $D \gg 5$ dimensional input spaces.  

To generate data, we begin by sampling the entries of a $d \times D$ projection matrix $P$ from a standard Gaussian distribution, and 
then project $n = 3000$ inputs $\bm{x}$ (of dimensionality $D \times 1$), with locations randomly sampled from a 
Gaussian distribution so that there is no input grid structure, into a $d=2$ dimensional space: $\bm{x}' = P \bm{x}$.
We then sample data $\bm{y}$ from a Gaussian process with an RBF kernel operating on the low dimensional inputs $\bm{x}'$. 
We repeat this process $30$ times for each of $D=1,2,3,4,5,6,7,8,9,10,15,20,25,\dots,100$.   

We now index the data $\bm{y}$ by the high dimensional inputs $\bm{x}$ and attempt to reconstruct the true low dimensional subspace 
described by $\bm{x}' = P\bm{x}$.  We learn the entries of $P$ jointly with covariance parameters through marginal likelihood optimisation (Eq.~\eqref{eqn: mlikeli}).  
Using a $(d=2)$ $50 \times 50$ Cartesian product grid for $2500$ total inducing points $U$, 
we reconstruct the projection matrix $P$, with the subspace error,
\begin{align}
\text{dist}(P_1,P_2)  = || G_1 - G_2 ||_2 \,,  \label{eqn: dist}
\end{align}
shown in Figure \ref{fig: proja}.  Here $G_i$ is the orthogonal projection onto the $d$-dimensional subspace spanned by the rows of $P_i$.   This metric is motivated by the 
one-to-one correspondence between subspaces and orthogonal projections.  It is bounded in $[0,1]$, where the maximum distance $1$ 
indicates that the subspaces are orthogonal to each other and the minimum $0$ is achieved if the subspaces are identical.
More information on this metric including how to compute $\text{dist}$ of Eq.~\eqref{eqn: dist} is available in chapter 2.5.3 of \citet{golub2012matrix}.  

We also make predictions on $n_* = 1000$ withheld test points, and compare i) our method using $\tilde{P}$ (GP Proj. Kernel), ii) an exact GP on the high dimensional space (GP Full), and iii) an exact GP on the true subspace (GP True), with results shown in Figure \ref{fig: projb}.   We average our results $30$ times for each value of $D$, and show $2$ standard errors.  The extremely low subspace and SMAE errors up to $D=40$ validate the consistency of our approach for reconstructing a ground truth projection.  Moreover, we see that our approach is competitive in SMAE (as defined in the previous section) with the best possible method, GP true, up to about $D = 40$.   Moreover, although the subspace error is moderate for $D=100$, we are still able to substantially outperform the standard baseline, an exact GP applied to the observed inputs $\bm{x}$.

We implement variants of our approach where $P$ is constrained to be 1) orthonormal, 2) to have unit scaling, e.g., $P_{\text{unit}} = \text{diag}(\sqrt{(PP^{\top})}^{-1}) P$.
Such constraints prevent degeneracies between $P$ and kernel hyperparameters from causing practical issues, such as length-scales growing to large values to shrink
the marginal likelihood log determinant complexity penalty, and then re-scaling $P$ to leave the marginal likelihood model fit term unaffected.  In practice we found that
unit scaling was sufficient to avoid such issues, and thus preferable to orthonormal $P$, which are more constrained.

\section{Discussion}
\label{sec: discussion}

We introduce massively scalable Gaussian processes (MSGP), which 
significantly extend KISS-GP, inducing point, and structure exploiting 
approaches, for:
(1) $\mathcal{O}(1)$ test predictions (section \ref{sec: fast}); 
(2) circulant log determinant approximations which (i) unify Toeplitz and Kronecker structure;
    (ii) enable extremely fast marginal likelihood evaluations (section \ref{sec: circulant}); 
    and (iii) extend KISS-GP and Toeplitz methods for scalable kernel learning in D=1 input dimensions,
    where one cannot exploit multidimensional Kronecker structure for scalability.
(3) more general BTTB structure, which enables fast exact multidimensional inference without 
requiring Kronecker (tensor) decompositions (section \ref{sec: BCCB}); and, 
(4) projections which enable KISS-GP to be used with structure exploiting
approaches for $D \gg 5$ input dimensions, and increase the expressive
power of covariance functions (section \ref{sec: project}).

We demonstrate these advantages, comparing to state of the art alternatives.  In particular, 
we show MSGP is exceptionally scalable in terms of training points $n$, inducing points $m$,
and number of testing points.  The ability to handle large $m$ will prove important for retaining
accuracy in scalable Gaussian process methods, and in enabling large scale kernel learning.

This document serves to report substantial developments regarding the SKI and KISS-GP
frameworks introduced in \citet{wilsonnickisch2015}.

\bibliographystyle{apalike}
\bibliography{mbibnew}

\clearpage{}\newpage{}
\appendix

\section{APPENDIX}
\label{sec: appendix}

\subsection{Derivatives For Normal Projections}
\begin{itemize}
\item apply chain rule to obtain from $\frac{\partial\psi}{\partial Q}$
from $\frac{\partial\psi}{\partial P}$
\item the orthonormal projection matrix $Q$ is defined by $Q=\text{diag}(\bm{p})P\in\mathbb{R}^{d\times D}$,
$\bm{p}=\sqrt{\text{diag}(PP^{\top})}^{-1}$ so that $\text{diag}(QQ^{\top})=\bm{1}$
\end{itemize}
\[
\frac{\partial\psi}{\partial Q}=\text{diag}(\bm{p})\frac{\partial\psi}{\partial P}-\text{diag}(\text{diag}(\frac{\partial\psi}{\partial P}P')\odot\bm{p}^{3})P
\]

\subsection{Derivatives For Orthonormal Projections}
\label{sec: ortho-project}

\begin{itemize}
\item the orthonormal projection matrix $Q$ is defined by 
  $Q=(PP^{\top})^{-\frac{1}{2}}P\in\mathbb{R}^{d\times D}$
  so that $QQ^{\top}=I$
\item apply chain rule to obtain from $\frac{\partial\psi}{\partial Q}$
  from $\frac{\partial\psi}{\partial P}$
\item use eigenvalue decomposition $PP^{\top}=VFV^{\top}$, define $S=V\text{diag}(\mathbf{s})V^{\top}$,
  $\mathbf{s}=\sqrt{\text{diag}(F)}$
\end{itemize}

\begin{eqnarray*}
\frac{\partial\psi}{\partial Q} & = & S^{-1}\frac{\partial\psi}{\partial P}-VAV^{\top}P\\
A & = & \frac{ V^{\top}S^{-1}
     (\frac{\partial\psi}{\partial P}P^{\top}+P\frac{\partial\psi}{\partial P}^{\top})
  S^{-1}V }
  {\mathbf{s}\mathbf{1}^{\top}+\mathbf{1}\mathbf{s}^{\top}}
\end{eqnarray*}

\begin{itemize}
  \item division in $A$ is component-wise
\end{itemize}

\subsection{Circulant Determinant Approximation Benchmark}

\begin{figure*}
\centering{}
\includegraphics[bb=45bp 15bp 880bp 590bp,clip,scale=.5]{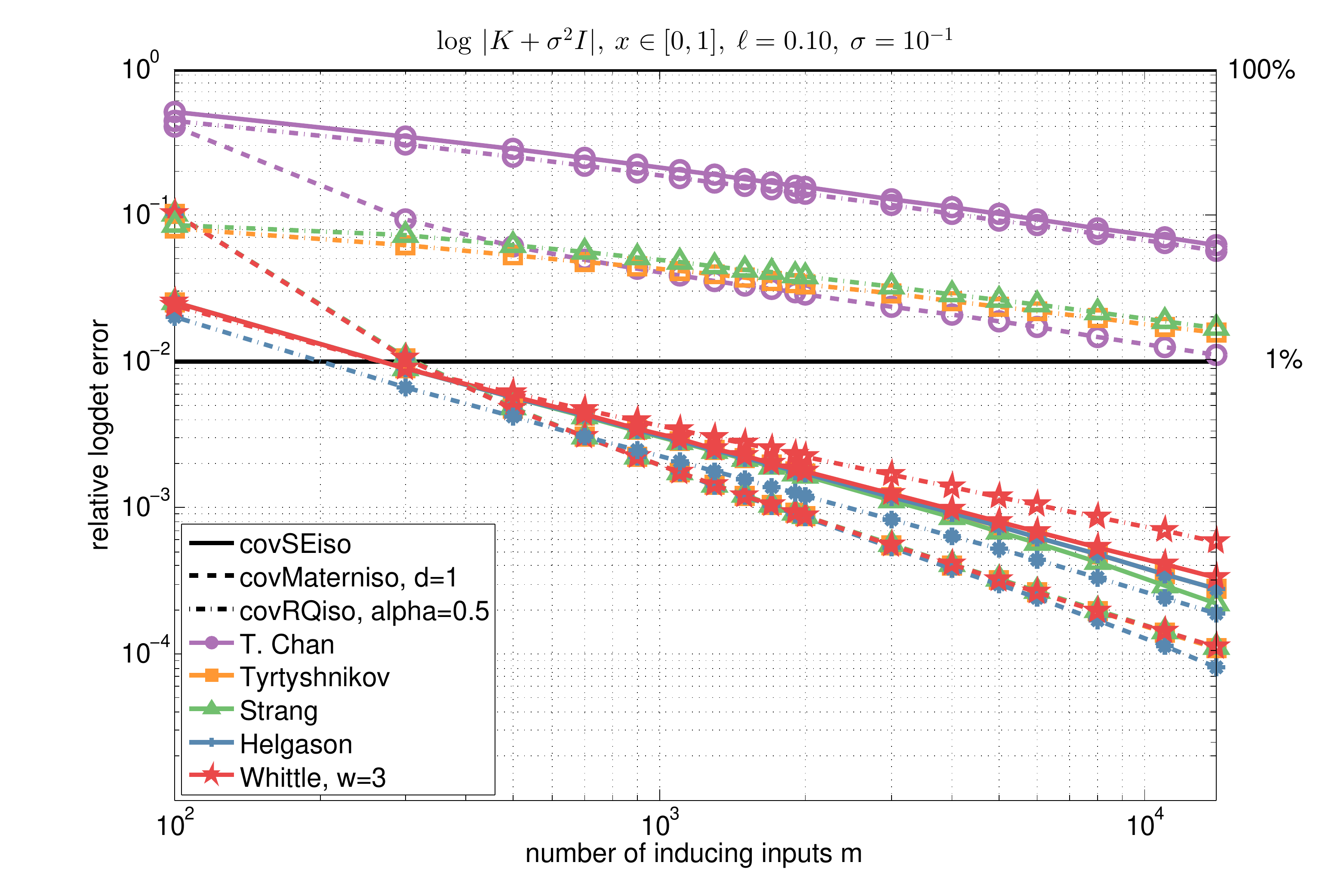} 
\includegraphics[bb=45bp 15bp 880bp 590bp,clip,scale=.5]{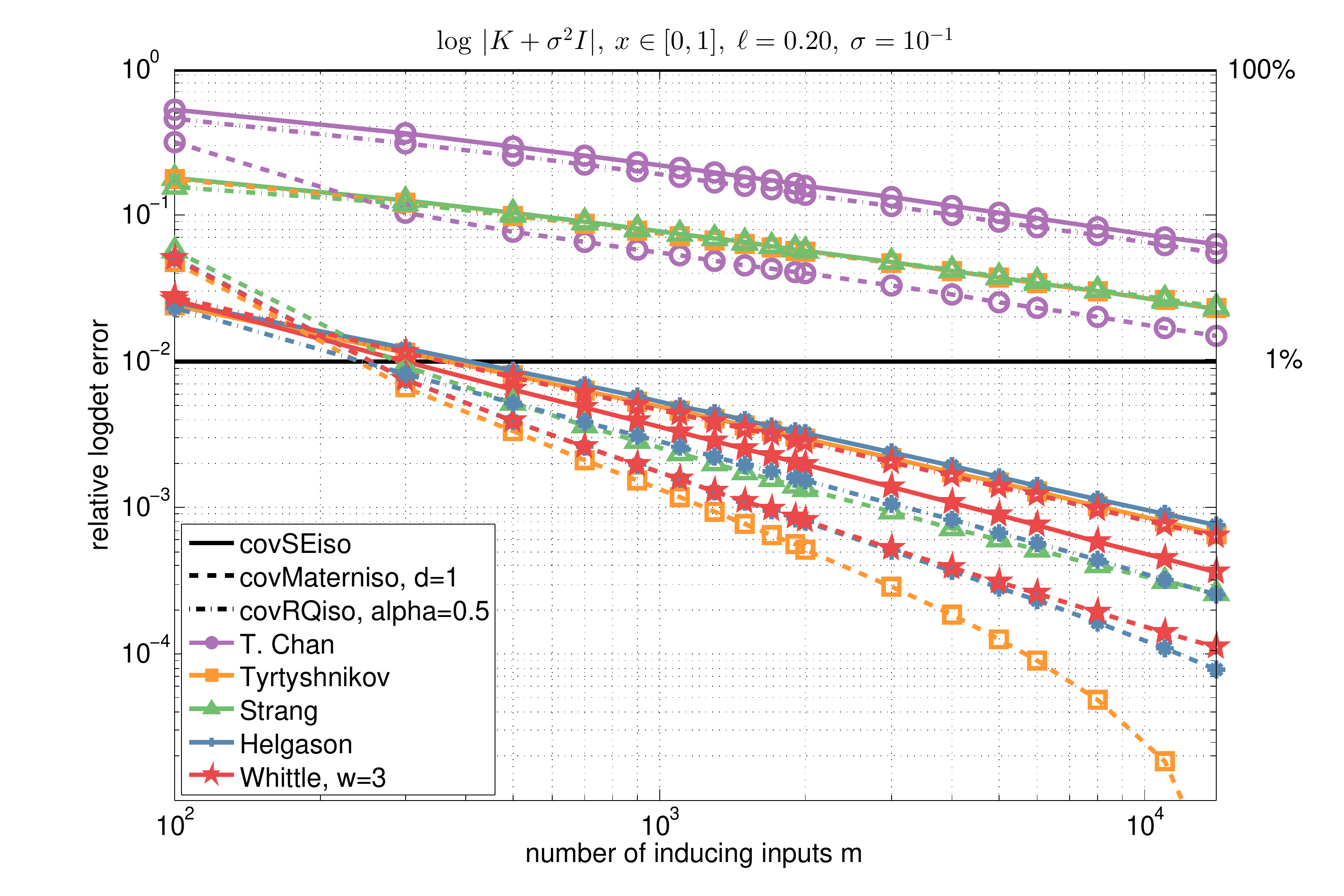} 
\caption{Additional benchmarks of different circulant approximations }
\end{figure*}

\begin{figure*}
\centering{}
\includegraphics[bb=45bp 15bp 880bp 590bp,clip,scale=.5]{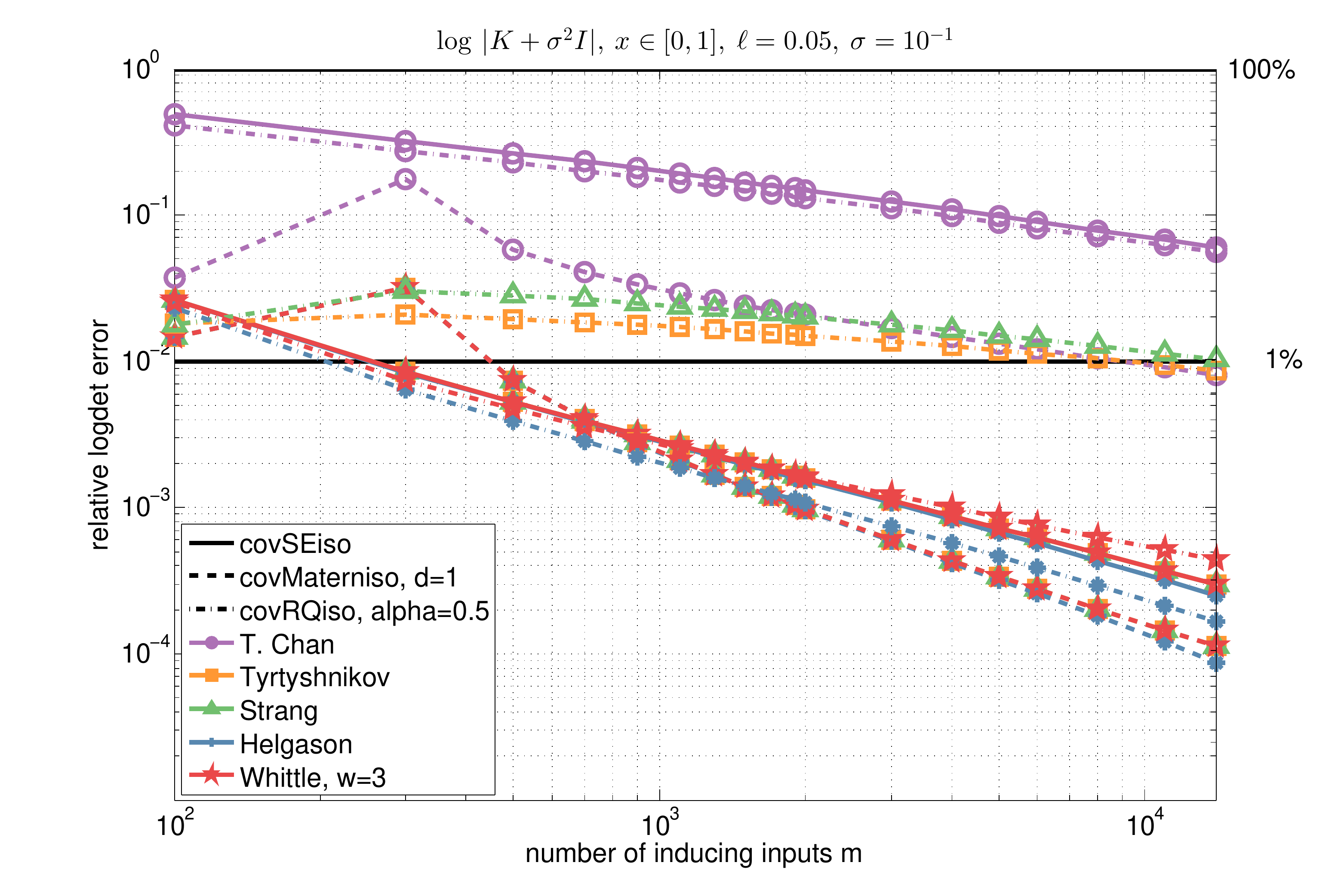} 
\includegraphics[bb=45bp 15bp 880bp 590bp,clip,scale=.5]{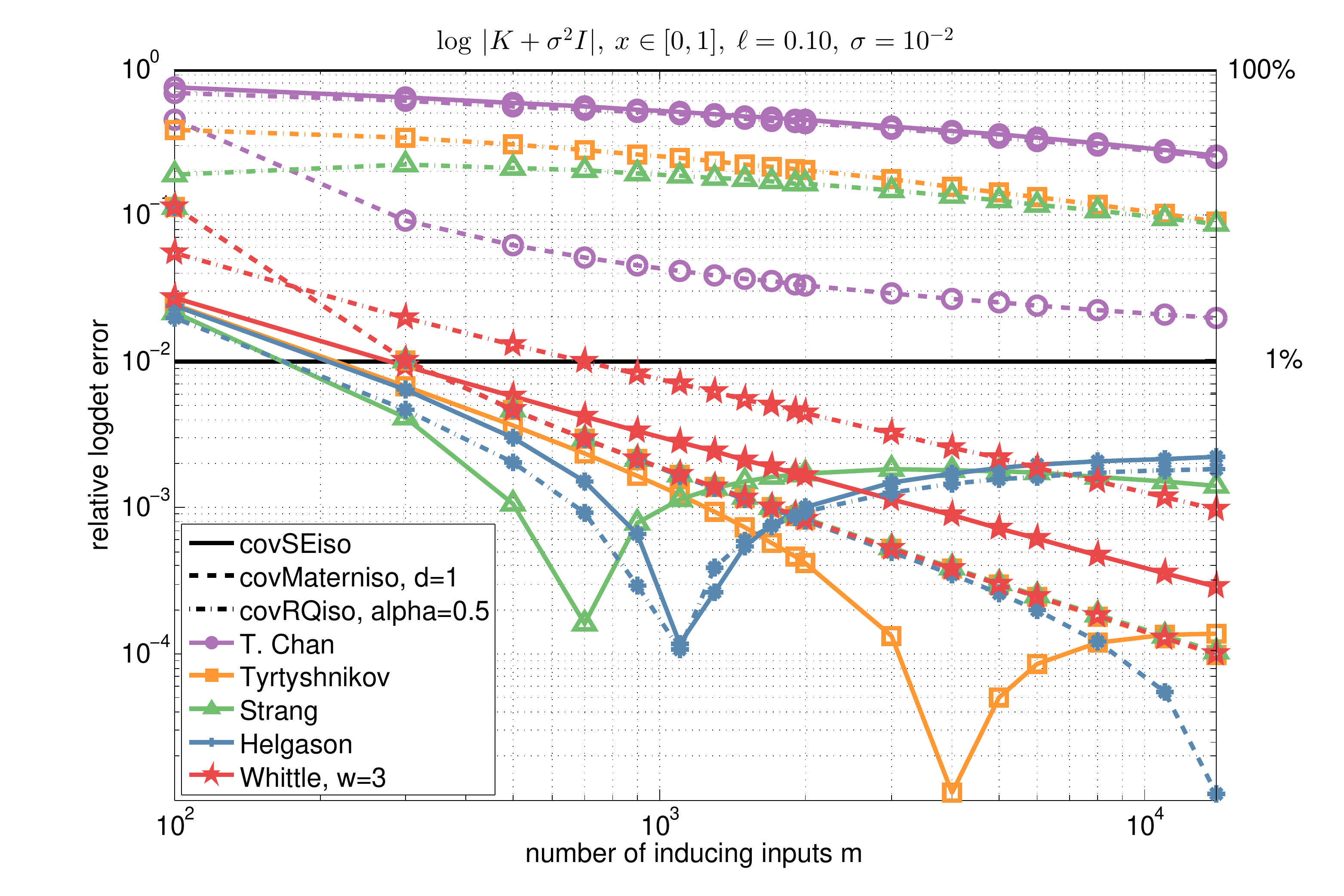}
\caption{Additional benchmarks of different circulant approximations }
\end{figure*}

\begin{figure*}
\centering{}
\includegraphics[bb=45bp 15bp 880bp 590bp,clip,scale=.5]{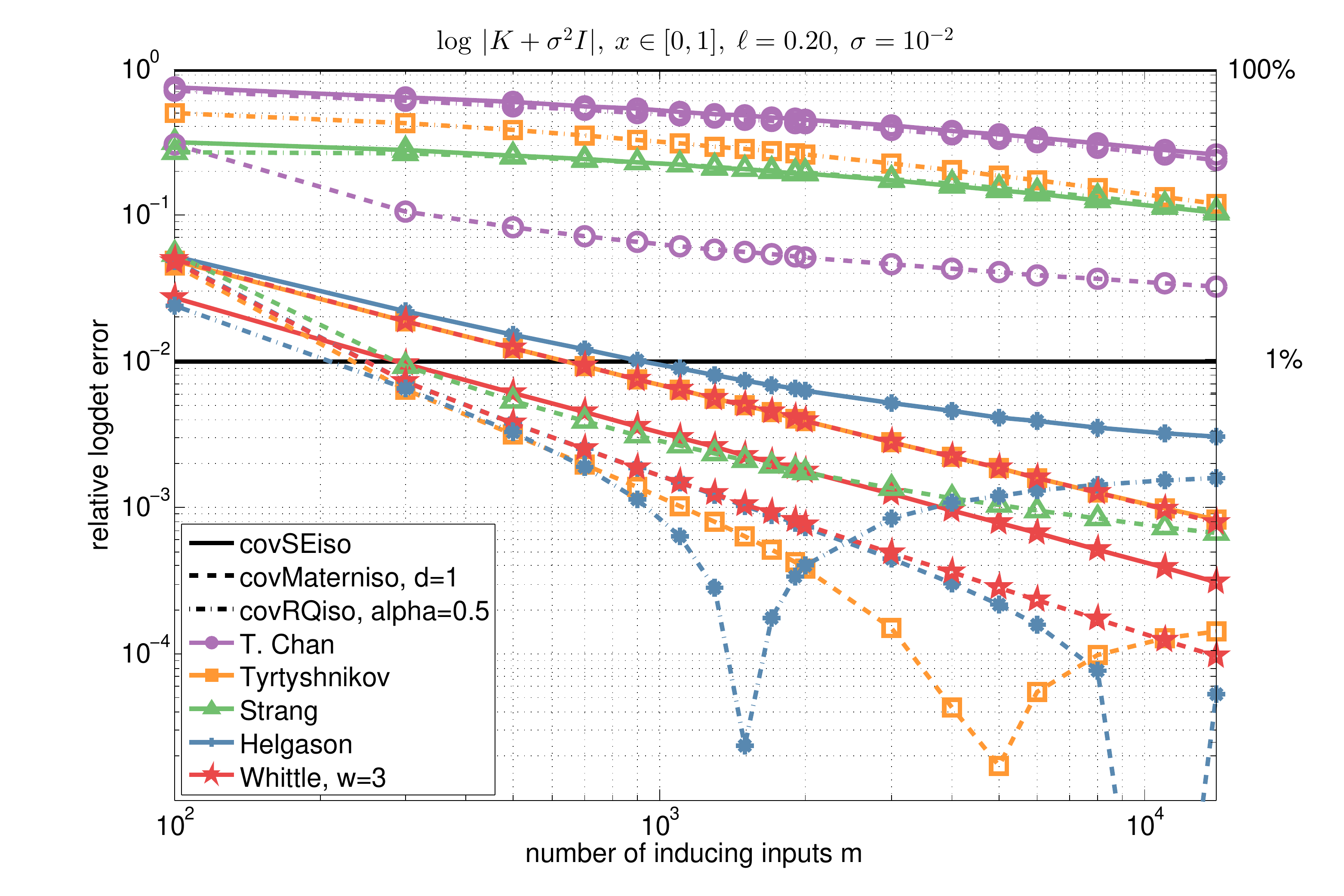}
\includegraphics[bb=45bp 15bp 880bp 590bp,clip,scale=.5]{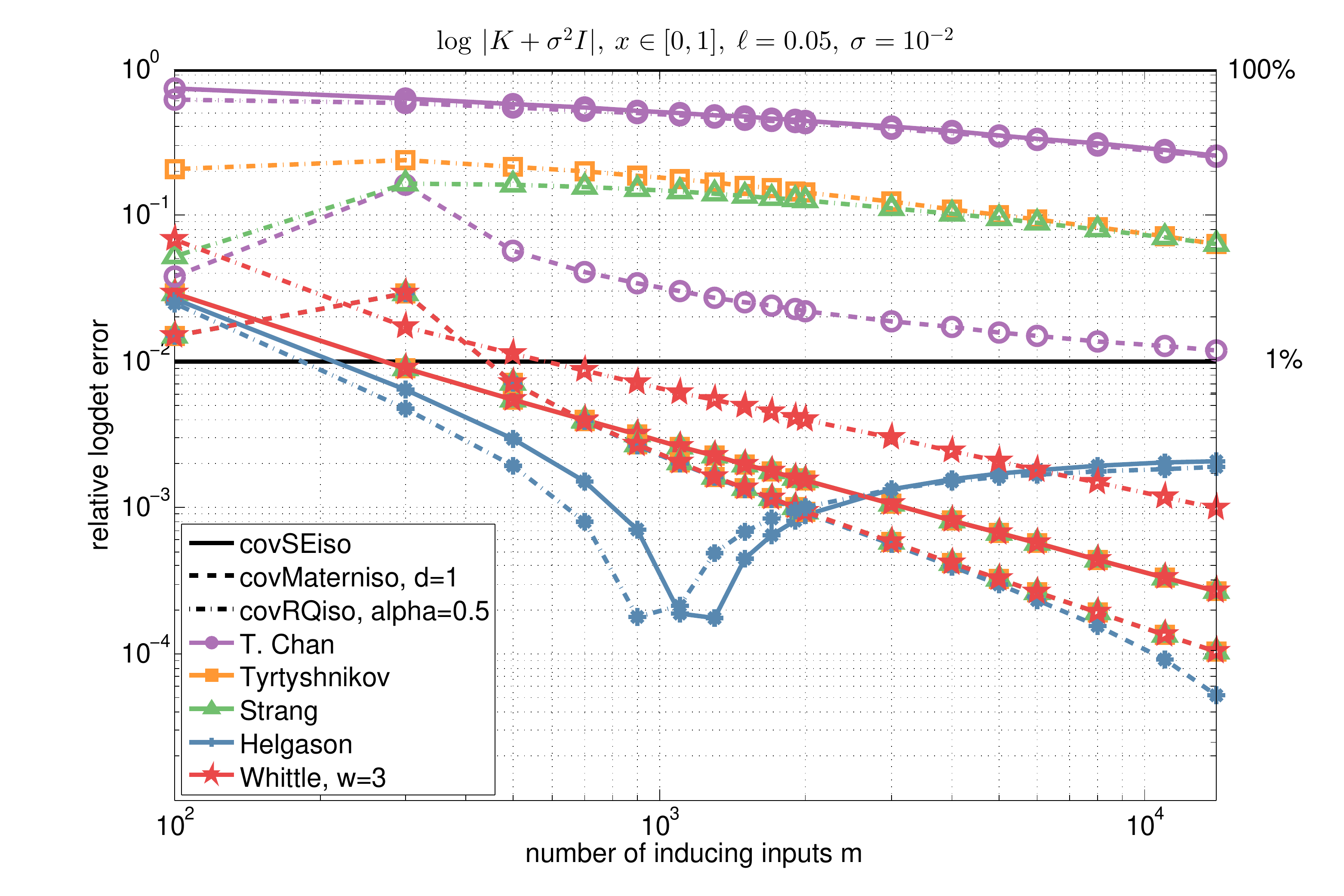}
\caption{Additional benchmarks of different circulant approximations }
\end{figure*}

\begin{figure*}
\centering{}
\includegraphics[bb=45bp 15bp 880bp 590bp,clip,scale=.5]{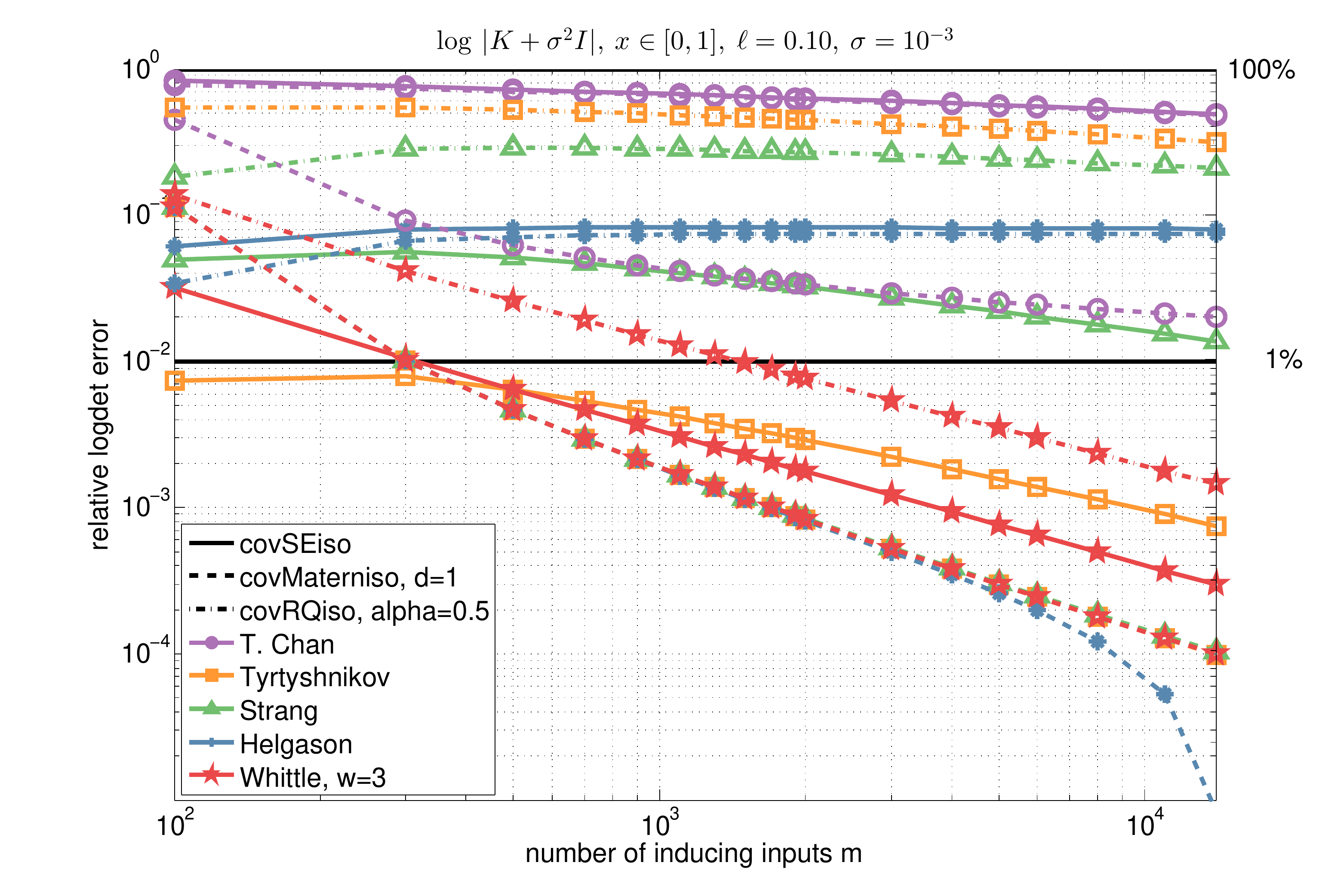}
\includegraphics[bb=45bp 15bp 880bp 590bp,clip,scale=.5]{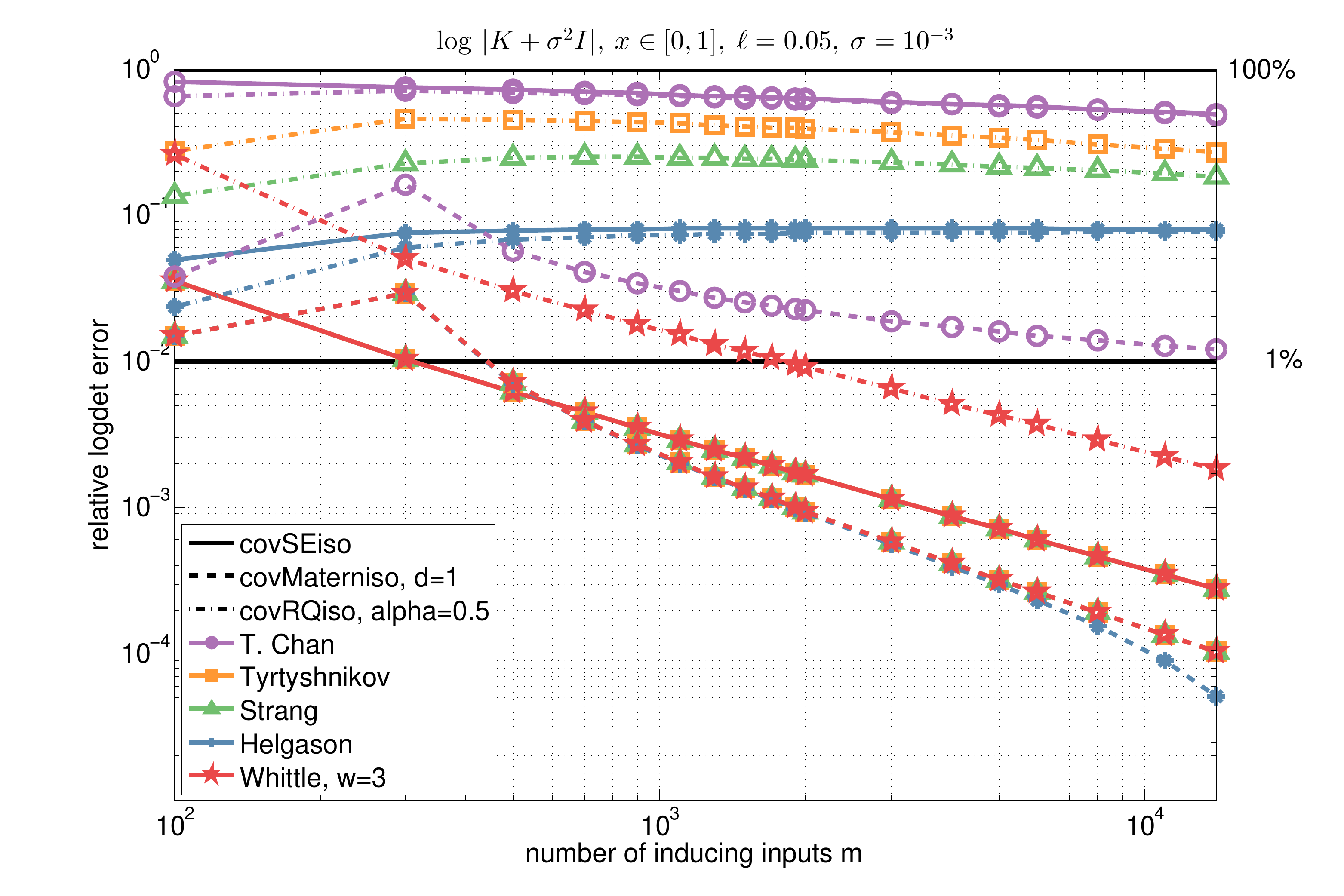}
\caption{Additional benchmarks of different circulant approximations }
\end{figure*}

\end{document}